\documentclass[sn-mathphys,Numbered]{sn-jnl}

\usepackage{multirow}
\usepackage{amsmath,amssymb,amsfonts}
\usepackage{amsthm}
\usepackage{mathrsfs}
\usepackage[title]{appendix}
\usepackage{xcolor}
\usepackage{textcomp}
\usepackage{manyfoot}
\usepackage{rotating}
\usepackage{booktabs}
\usepackage{caption}
\usepackage{graphicx}

\title[CACTUS]{CACTUS: a Comprehensive Abstraction and Classification Tool for Uncovering Structures}
\begin{document}


\author*[1]{\fnm{Luca} \sur{Gherardini}}\email{l.gherardini@sanoscience.org}

\author[1]{\fnm{Varun Ravi} \sur{Varma}}\email{v.varma@sanoscience.org}

\author[1]{\fnm{Karol} \sur{Capała}}\email{k.capala@sanoscience.org}

\author[2]{\fnm{Roger} \sur{Woods}}\email{r.woods@qub.ac.uk}

\author[1]{\fnm{Jose} \sur{Sousa}}\email{j.sousa@sanoscience.org}

\affil[1]{\orgdiv{Personal Health Data Science Team}, \orgname{Sano Centre for Personalised Computational Personalised Medicine}, \orgaddress{\street{Czarnowiejska 36 building C5}, \city{Krakow}, \postcode{30-054}, \state{Poland}}}

\affil[2]{\orgdiv{School of Electronics, Electrical Engineering, and Computer Science}, \orgname{Queen's University Belfast}, \orgaddress{\street{6 Malone Road}, \city{Belfast}, \postcode{BT9 5BN}, \state{Northern Ireland, United Kingdom}}}

\abstract{The availability of large data sets is providing an impetus for driving current artificial intelligent developments.
There are, however, challenges for developing solutions with small
data sets due to practical and cost-effective deployment and the opacity of deep learning models. 
The Comprehensive Abstraction and Classification Tool for Uncovering Structures called CACTUS is presented for improved secure analytics by effectively employing explainable artificial intelligence. 
It provides additional support for categorical attributes, preserving their original meaning, optimising memory usage, and speeding up the computation through parallelisation. 
It shows to the user the frequency of the attributes in each class and ranks them by their discriminative power.
Its performance is assessed by application to the Wisconsin diagnostic breast cancer and Thyroid0387 data sets.
}


\maketitle

\section{Introduction}\label{sec1}
New computational capabilities and high volumes of data have driven the development and adoption of new artificial intelligence (AI) algorithms.
The main protagonist of the machine learning (ML) domain has undoubtedly been deep learning (DL) due to its impressive performance in many sectors. 
DL technologies are now being widely used to automate tasks with results substantially comparable to humans.
However, challenges exist for the sustainability of deep learning approaches: the need to train these models with the large data sets needed requires extreme resources in terms of storage and computational power, costing millions~\cite{Li2023, Anthony2020, Selvan2022, Strubell2020}; their extreme hunger for data makes these models hard to apply on small data sets due to practical and cost-effective deployment; the opacity of the ``black box'' DL models makes them hard to trust, especially in sensitive fields like medicine.
This contrasts with the ``right to explainability'' ambition defined in the recital 71 of the general data protection regulation (GDPR) of the European Union and in the algorithmic accountability act decreed by the US congress (H.R. 6580 of 2022)~\cite{MacCarthy2020, Goodman2016}.

The rising interest in explainable AI (XAI) is formulated in terms of \emph{interpretability} and \emph{explainability}~\cite{Guidotti2021}.
Based on multi-agent systems, interpretability is defined as the process of assigning a subjective meaning to something, while explainability is concerned with transforming an object in something more easily or effectively interpretable~\cite{Ciatto2020}.
Some methods act to achieve this explainability by attempting to peer into the black box~\cite{Handelman2019} through post-hoc models~\cite{Li2022}, trying to produce explanations by applying explainable-by-design models such as decision trees, to neural networks to approximate their behaviour and prove their \emph{robustness} either locally or globally.
Angelov et al.~\cite{Angelov2020} proposed new DL architectures to augment their explainability or performed meta-analyses on their functioning~\cite{Angelov2021}.
The trade-off between explainability and accuracy is widely discussed~\cite{London2019}, but one point in favour of emphasising the former is the greater improvement and testing potential of methods that are easier to understand.

In our previous work on a \emph{small and iNcomplete dataset analyser} (SaNDA)~\cite{SaNDA}, we tackled these problems using abstractions to enhance privacy, in order to allow statistical approaches to be deployed, share anonymous data among different research centres, and construct knowledge graphs to represent the information extracted from the data.
A key contribution of this work is that we use abstractions to partition the set of values of an attribute in \emph{Up} and \emph{Down} flips to represent them in a more intuitive, probabilistic, and anonymous format.
In this paper, we further exploit this concept to create the Comprehensive Abstraction and Classification Tool for Uncovering Structures (CACTUS), which supports additional flips for categorical attributes, preserving their original meaning, by optimising its memory usage, and by speeding up the computation through parallelisation.
The architecture of CACTUS incorporates the original concept of SaNDA but in a broader manner, allowing it to automatically stratify and binarise for multiple configurations at once. This allows the creation of binary decision trees and correlation matrices along the abstractions, and by storing intermediate results for easy revision and integration with other tools.
In addition, the output structure of SaNDA has been improved and rationalised to transform it into a research tool that can be easily and broadly used by the scientific community.

In order to demonstrate and evaluate the work, we use the Wisconsin diagnostic breast cancer data set (WDBC, available at \url{archive.ics.uci.edu/ml/machine-learning-databases/breast-cancer-wisconsin/})~\cite{WinsconsinBreastCancer,Wolberg1995Breast} and the Thyroid0387 data sets~\cite{Thyroid_dataset} to gauge and to show the performance and ease of use of CACTUS.
The work makes a number of contributions:
\begin{itemize}
    \item A faster and lighter data abstraction as well as a mechanism for knowledge graph creation is presented.
    \item The approach naturally deals with categorical features.
    \item The creation of additional metrics to interpret the classification process.
    \item Auxiliary analyses of the data set to identify patterns and contrast them to the insights provided by the classification process.
\end{itemize}
The paper is organised as follows; Section 2 covers the mechanism used for abstracting values in SaNDA. We then show how CACTUS abstracts and extracts information from the WDBC and the Thyroid0387 data sets in Section 3. In Section 4, it is shown how we operated a meta-analysis to interpret its internal reasoning and compare the approach to existing literature.

\section{Methods}
CACTUS has been implemented in Python3~\cite{Python3}, using scientific libraries such as Numpy~\cite{Numpy} and pandas~\cite{Pandas, mckinney-proc-scipy-2010} to achieve high performance while remaining easy to debug. 
It offers various options to the user for handling a given data set, such as replacing values, dropping certain columns, and considering what value is a not-a-number (NaN). This is subsequently encapsulated in a Yaml configuration file, thereby allowing the computation to be customised and different analyses to be performed.
It can be compiled in Cython to obtain a faster version that uses a command line tool and offering a convenient way to switch from multi-class to binary classification by specifying how to \emph{binarise} the label to predict.
Both the binarisation and the stratification functionality can be specified multiple times and switched in a seamless manner.
Attributes for stratification have to be integer and must have less than 10 unique values.

The pipeline implemented in CACTUS is organised into three independent functional modules: preprocessing, abstraction, and correlation.

The preprocessing module computes the binary decision trees.
It uses scikit-learn~\cite{Scikit-learn} to assess the combinations of columns which represent the most reasonable approach to compute the correlation.
Performing the preprocessing can speed up the correlation phase, if the data set presents particular patterns of missing values, and it may make the correlation matrix less sparse.

The abstraction module is similar to the one implemented by SaNDA~\cite{SaNDA}, but its receiver operating characteristic (ROC) curve has been adapted to categorical values to preserve their original meaning.
An attribute is recognised automatically if it has less than 10 unique values, but can be forced to be categorical by casting its values, allowing the system to make categorical an attribute that would not spontaneously fit into the definition.
This can be useful for attributes that assume few values, either for high percentages of NaNs or for the specific population that is being analysed; thus giving a finer granularity, or allowing a stratification by those values.

Let $A$ be a categorical attribute assuming values ${1, 2, 3}$; its states will be coded in the \emph{flips} $A_1, A_2, A_3$, each with its own significance and connections with other \emph{flips}.
In SaNDA, all attributes were simply labeled as Up and Down, and thus their semantic meaning (i.e. Genetic variables) was lost.
The graph-tool library~\cite{Graph-tool} used for constructing the knowledge graphs has been kept, but the stochastic community detection algorithm~\cite{Markov_Chain_Monte_Carlo_Stochastic_Partitioning} has been replaced by the greedy~\cite{Greedy_modularity_communities}, the label propagation~\cite{Asyn_lpa_communities}, and the Louvain~\cite{Louvain_communities} deterministic approaches offered by the networkX library~\cite{NetworkX}.

Previously in SaNDA~\cite{SaNDA}, adjacency was stored and computed using adjacency lists, which were very cumbersome particularly for a large data set.
We removed this limitation by computing these connections on-the-fly as required, saving disk and memory space at a reduced computational cost.
This has a minimal impact on performance because the connections are computed once for each graph in parallel, and the connections are saved in a comma-separated value (CSV) file in a much more convenient way than adjacency lists.
Such structures can be evaluated using the notions of modularity, coverage, and performance~\cite{Greedy_modularity_communities, Partition_quality} and automatically computed, compared and stored during the process.
The graphs are stored in graphml format, which allows for additional analyses using different software, like Gephi~\cite{Gephi}.

After building the knowledge graphs, CACTUS classifies each record individually and compares the two available classification means, PageRank and Probabilistic~\cite{SaNDA}, against the ground truth.
Having two classification mediums is useful in identifying what is more important to predict the target value between the probability of an attribute and its centrality for the interaction networks.
Also, having similar performance between PageRank and Probabilistic provides a validation of the knowledge graph, because the connections modify the significance of each marker and drive the former classification method.

Another addition to SaNDA is computing balanced accuracy for multi-class classification, as described below,
\begin{equation}
    \frac{1}{N} \sum_{c=1}^N \frac{\omega_c}{|E_c|}
    \label{eq-bal-acc}
\end{equation}
where $N$ is the total number of classes, $\omega_c$ are the number of correctly classified instances, and $E_c$ is the set containing all the instances of class $c$.
CACTUS automatically plots how the probability distribution of each marker changes across the classes, giving insights on the variables driving the classification towards particular classes.
Through Equation \eqref{eq-rank}, we model the rank $R_{x_f}$ for each flip $f$ of a marker $x$ by considering how the conditional probability $P(x|S_i)$ of the flip $x$ given the class $S_i$ changes across the $N$ considered classes.

\begin{equation}
    R_{x_f} = \sum_{i=1, j>i}^{N} \frac{|P(x_f|S_{i-1}) - P(x_f|S_j)|}{_NC_2}
    \label{eq-rank}
\end{equation}

By considering the unordered pairs of states $(S_i, S_j)$ do not assume the classes to be ordered, and can average them by the number of pairs obtained, which is given by the combination $_NC_2$, we can average the rank over the $F$ flips of the marker $x$. Thus we get Equation \eqref{eq-avg-rank}, 

\begin{equation}
    \Bar{R}_x = \frac{1}{F}\sum_{f=0}^{F} R_{x_f}.
    \label{eq-avg-rank}
\end{equation}

If enabled, the correlation module computes the correlation for the set of columns kept by the preprocessing; otherwise, it is calculated for the whole data set.
In both cases, the original value, not the abstracted data, are used.
A warning is issued if attribute combinations have a correlation of $\pm1$, because they might indicate either redundant or very interesting pairs of features.
A correlation graph is also computed, using the graph-tool~\cite{Graph-tool}, and the PageRank and Laplacian scores are computed along the same community detection algorithms applied in the knowledge graphs.
The self-connections can be removed optionally from the correlation graph, as they will always correspond to 1, and the graph will be saved in the graphml format as well.
A minimum spanning tree is computed on the correlation graph to show what attributes are better connected in the graph.
All these modules produce intermediate outputs that can be integrated, accessed or elaborated differently from what is automatically done.

The WDBC data set contains the features of digitalised images of fine needle aspirates of breast mass for a total of 569 patients.
Enquiries about the exact description of the features and of the whole data set are readdressed to its original documentation on \url{archive.ics.uci.edu/ml/machine-learning-databases/breast-cancer-wisconsin/wdbc.names}.
It naturally presents a binary classification scenario, where breast cancer is labelled as benign or malignant and characterised by 30 real-value attributes.
These attributes can either be actual measures, or their standard error and worst value.
None of the attributes presents a NaN, and we did not artificially introduce them as this has already been considered and discussed in our previous work~\cite{SaNDA}.

The Thyroid data set consists of 9172 records composed of 29 attributes, 21 of which are binary, 7 are continuous and the last one is the referral source.
The \emph{Diagnosis} column consists of a string containing several information on the conditions of the patient, namely hyperthyroid and hypothyroid conditions, increased/decreased binding protein, general health, and treatments.
To perform a classification task, we considered patients with a "concurrent non-thyroidal illness" (K), hyperthyroid (letters A, B, C, D), or hypothyroid (letters E, F, G, H) conditions, and we re-code them in a multiple classification task between healthy (0), hyperthyroid (1) and hypothyroid (2) patients, as the previous experiments listed in the data set description.
This shrank the amount of records to 1371.
The Thyroid data set also contains dual diagnosis in the form $X|Y$, which is interpreted as "consistent to $X$, but $Y$ is more likely".
Therefore, when this ambiguity was present, we assigned $Y$ if it was one of the considered letters (A, B, C, D, E, F, G, H, K), else we assigned $X$ if $X$ was valid, otherwise we ignored the record.

\section{Results}

\subsection{WDBC data set}

The results of the classification of the WDBC data set are shown in Table~\ref{tab:wdbc-classification}.
Compared with the same analysis operated for SaNDA~\cite{SaNDA}, CACTUS maintains consistent performance.
In Figure~\ref{fig:wdbc-all-node-significance}, the probabilities of each marker across the benign (0) and malignant (1) breast cancers allow to compute the importance of each variable in discriminating across the available classes by applying Equation \eqref{eq-avg-rank}. 

\begin{figure}[p]
	\centering
	\includegraphics[width=\textwidth]{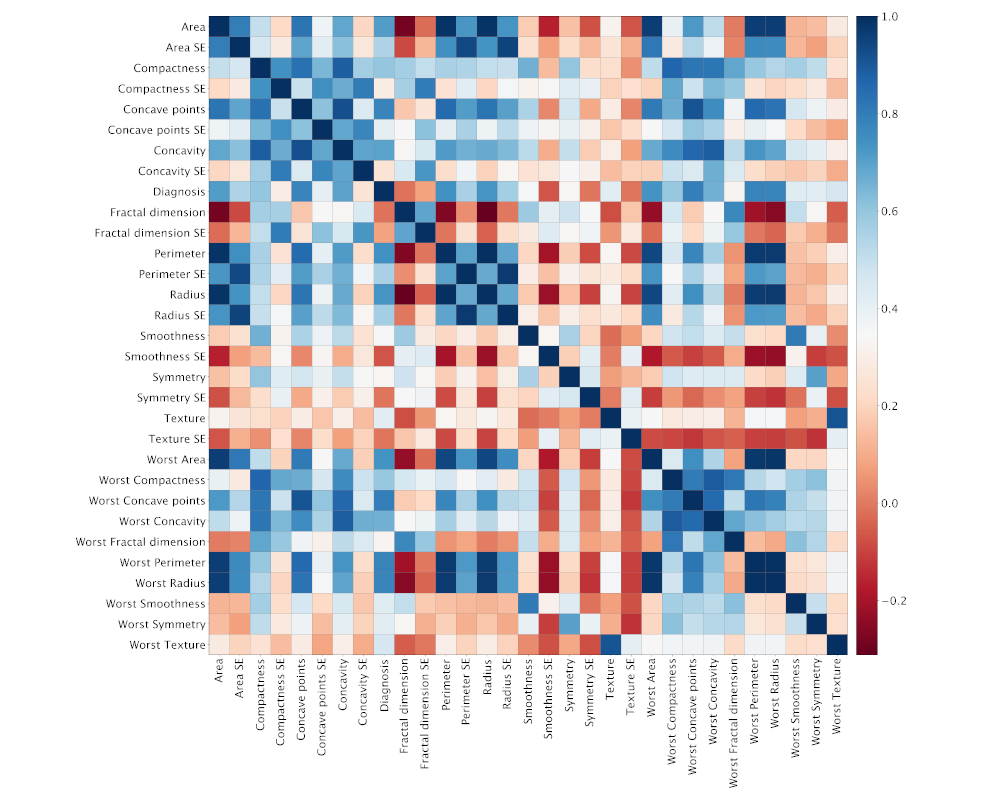}
	\caption{Correlation matrix between the flips of the WDBC.}
	\label{fig:wdbc-correlation-matrix}
\end{figure}

\begin{figure}[p]
	\centering
	\includegraphics[width=\textwidth]{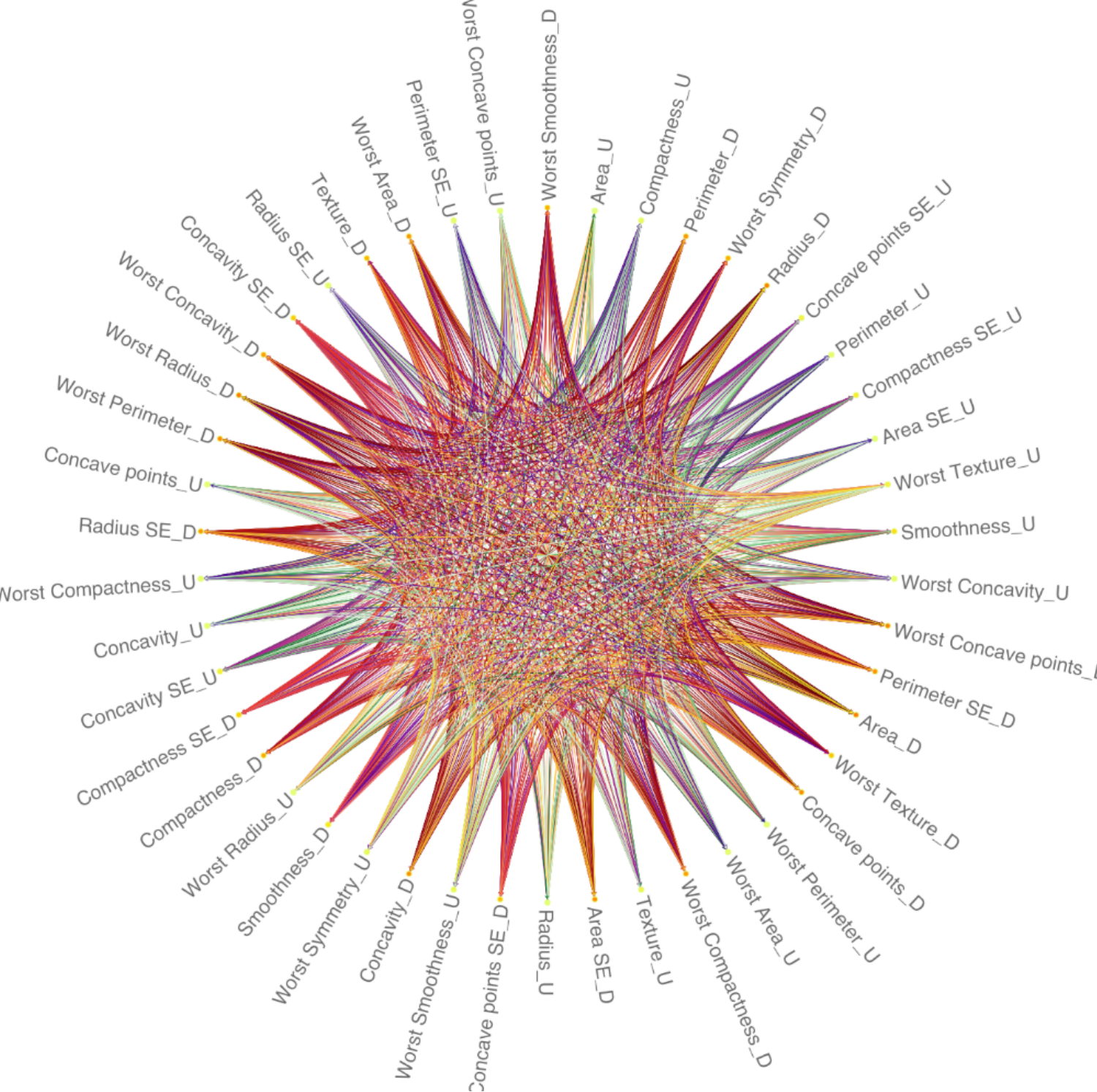}
	\includegraphics[width=.7\textwidth]{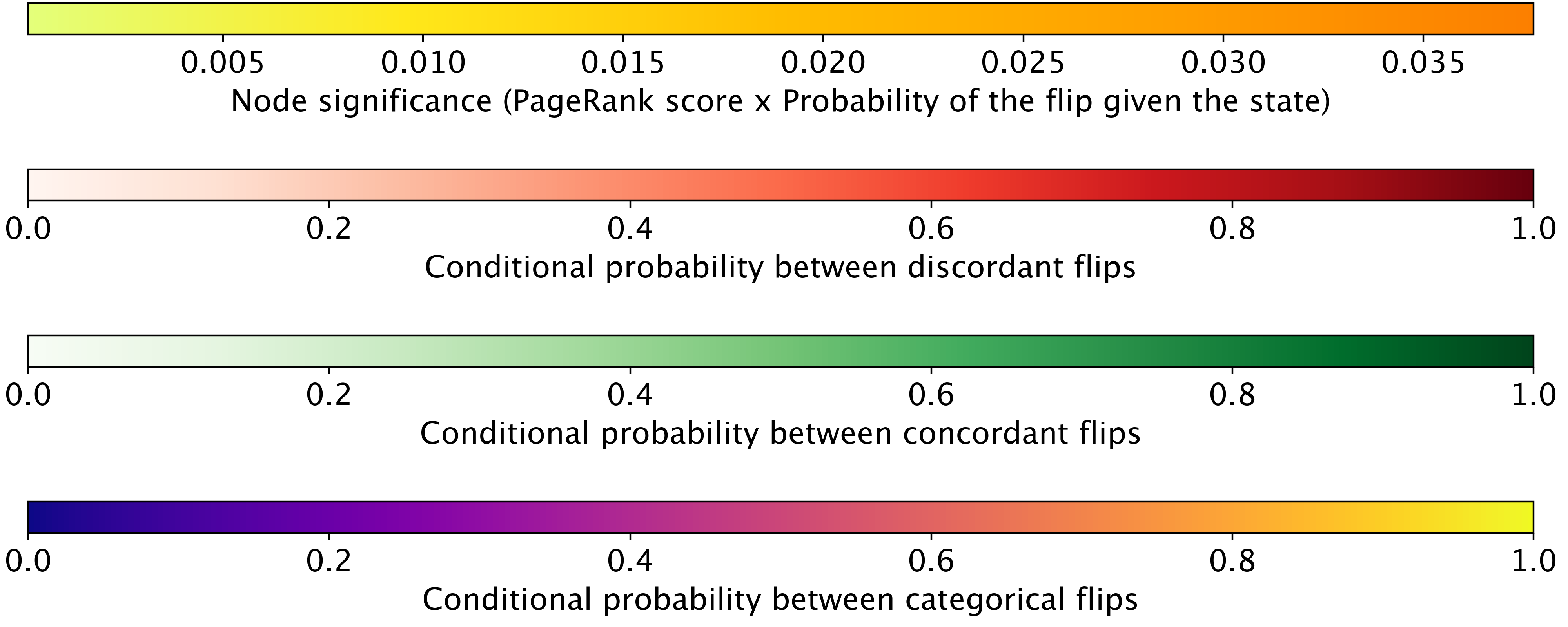}
	\caption{Knowledge graph for benign breast cancers. The node colours denote the corrected PageRank significance, while the edges are colored in different palettes depending on the kind of flips they are linking: green for \emph{Up} flips, red for \emph{Down} flips, and \emph{plasma} for \emph{categorical} flips.}
	\label{fig:wdbc-graph_0}
\end{figure}

\begin{figure}[p]
	\centering
	\includegraphics[width=\textwidth]{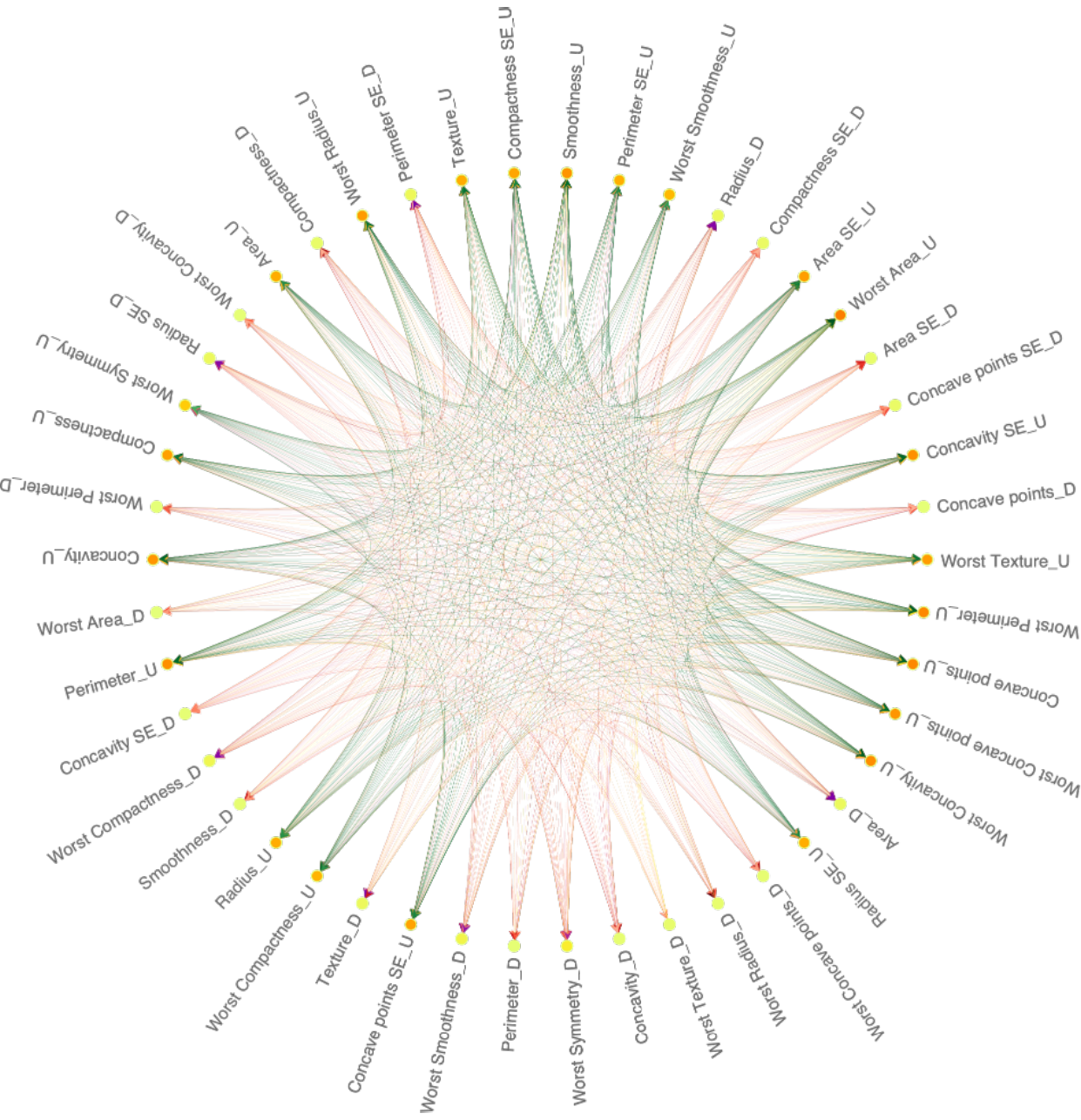}
	\includegraphics[width=.7\textwidth]{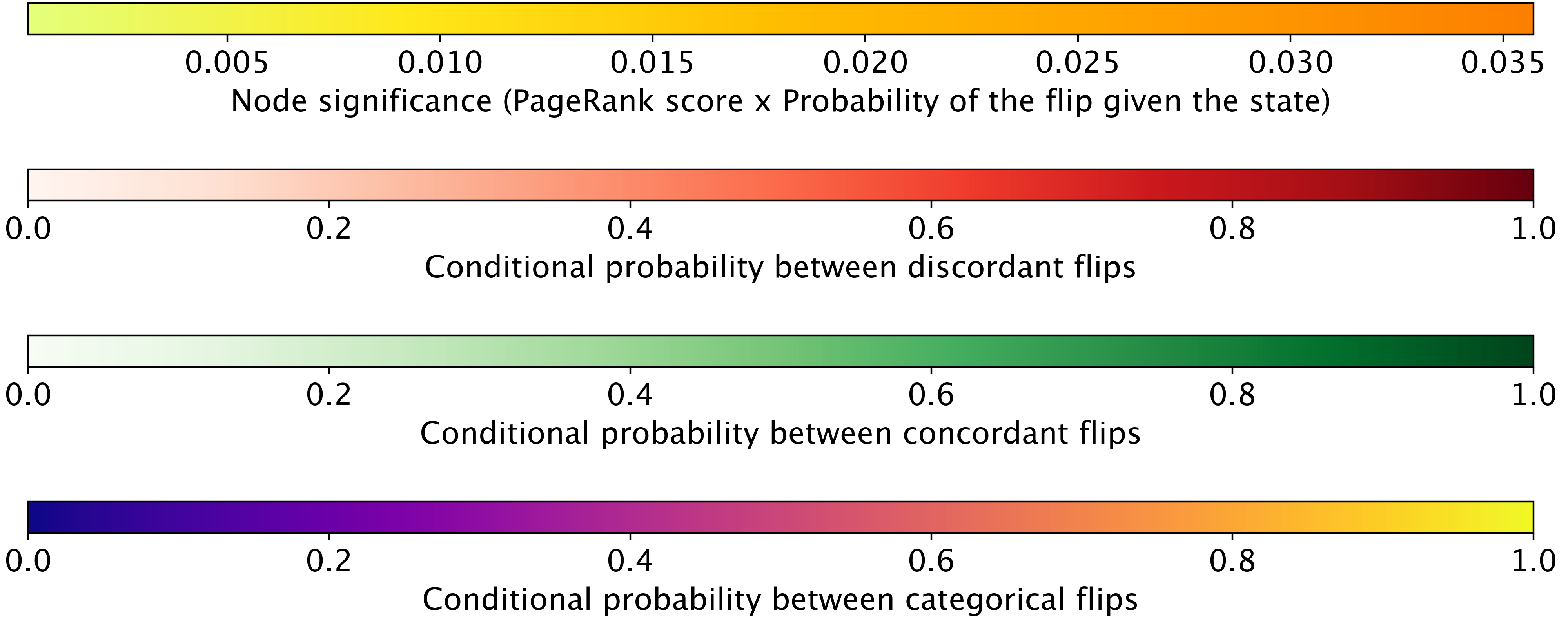}
	\caption{Knowledge graph for malignant breast cancers. The node colours denote the corrected PageRank significance, while the edges are colored in different palettes depending on the kind of flips they are linking: green for \emph{Up} flips, red for \emph{Down} flips, and \emph{plasma} for \emph{categorical} flips.}
	\label{fig:wdbc-graph_1}
\end{figure}

\begin{figure}[p]
	\includegraphics[width=.5\textwidth]{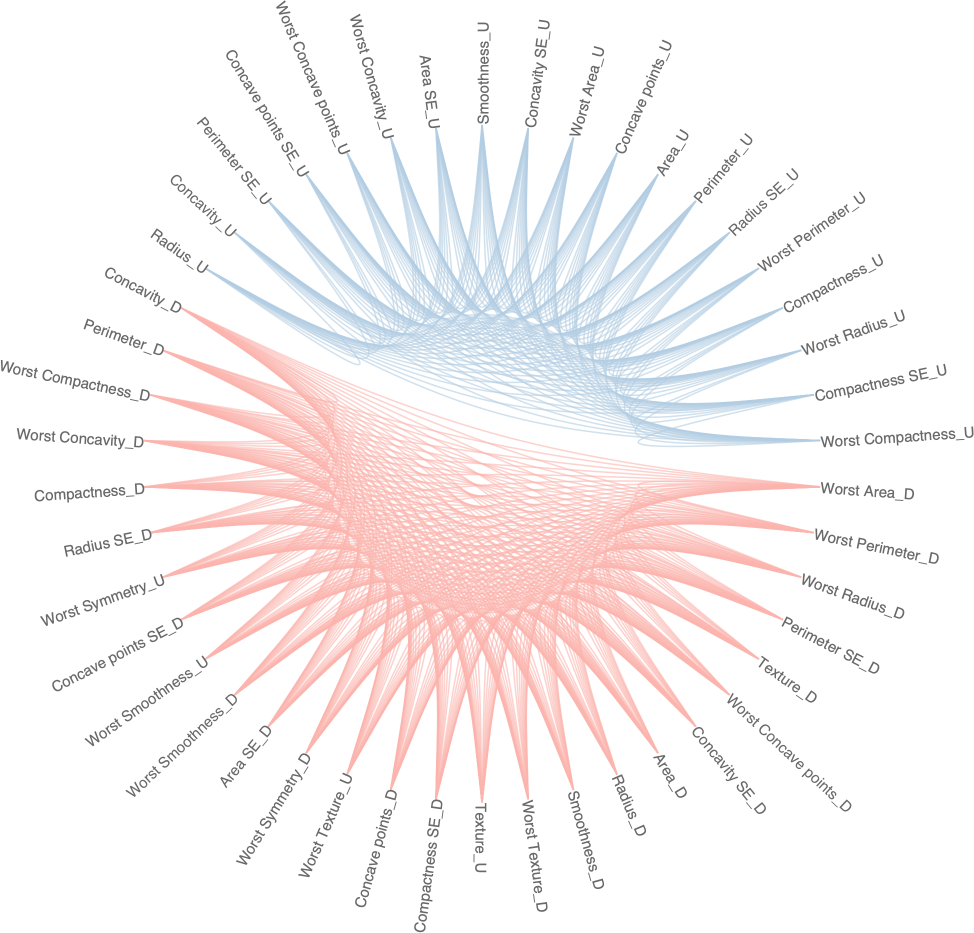}
	\includegraphics[width=.5\textwidth]{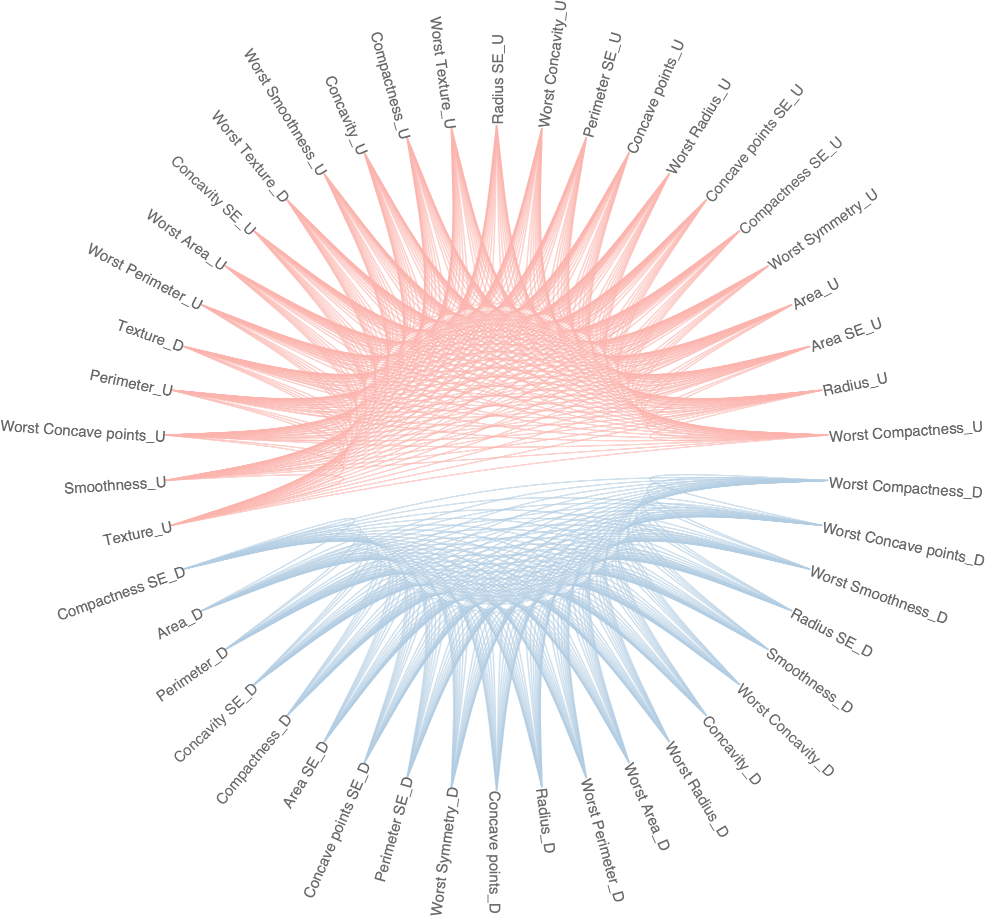}
	\caption{Communities obtained through the Greedy algorithm in benign (left) and malignant (right) breast cancers. The blue partition in the benign graph is fully contained in the red partition of the malignant but the latter comprises additional flips. Specularly, the blue community in the malignant graph is contained in the red community in the benign graph. The members changing membership result from the dynamics distinguishing malignant and benign breast cancer.}
	\label{fig:wdbc-greedy-communities}
\end{figure}

\begin{figure}[p]
    \includegraphics[width=.5\textwidth]{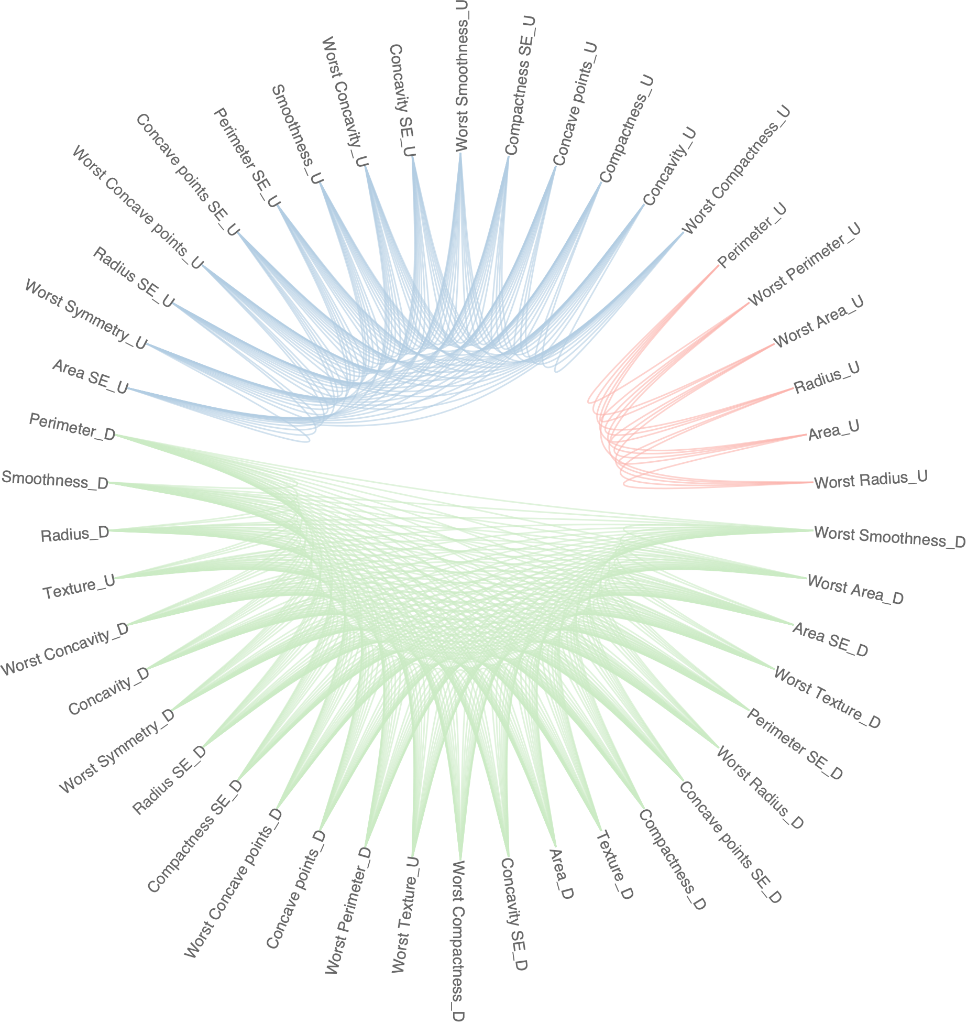}
    \includegraphics[width=.5\textwidth]{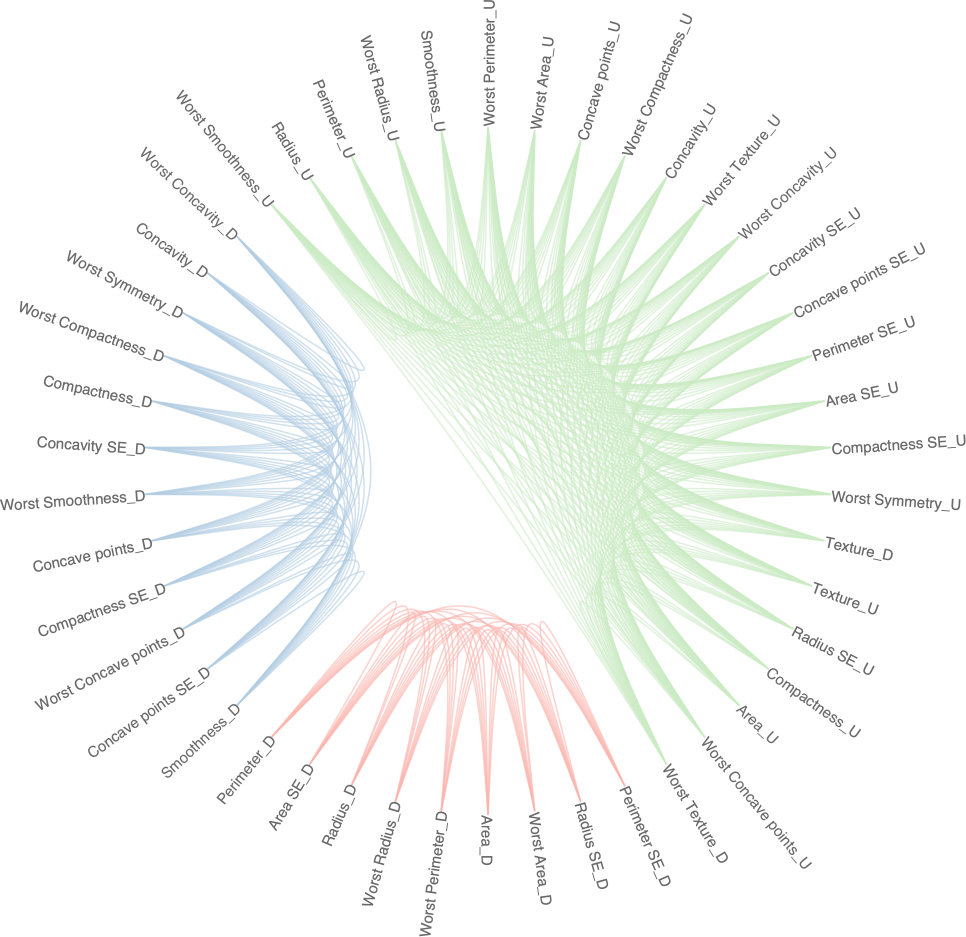}
    \caption{Communities obtained through the Louvain algorithm in benign (left) and malignant (right) breast cancers. Communities with the same colour in benign and malignant breast cancers have a subset of specular flips, for example \emph{Area\_U} in the former and \emph{Area\_D} in the latter red cluster, showing a change in the ``polarity'' of the markers in the same structure.}
    \label{fig:wdbc-louvain-communities}
\end{figure}

Figures~\ref{fig:wdbc-graph_0} and~\ref{fig:wdbc-graph_1}, indicate the strongest kind of connection changes in the knowledge graphs for the benign (red links) and the malignant breast cancers (green links), respectively, showing how discordant flips are tied more tightly in the former than in the latter. 
Also, the nodes where the strongest connections are converging (darker node colour) are changing. 
In Figures~\ref{fig:wdbc-greedy-communities} and~\ref{fig:wdbc-louvain-communities}, the difference across the Greedy and the Louvain algorithms are shown (Label propagation has been omitted because, for this data set, it returned a single community).
They show, respectively, how communities have a strong subset resistant to changes and a subset more weakly bound and inclined to change membership across cancer typology, and how specular flips can have the same role in the two knowledge graphs, forming the same structure.
Figure~\ref{fig:wdbc-correlation-matrix} shows the correlation among the flips, plus the \emph{Diagnosis} column.

\begin{table}[htb]
	\begin{tabular}{cc|cc}
		& Accuracy & Balanced Accuracy \\ \hline
		CACTUS & PageRank      & $0.9508$   & $0.9493$           \\
		& Probabilistic & $0.9543$   & \textbf{0.954} \\
    \hline
        SaNDA~\cite{SaNDA} & PageRank & 0.9467 & 0.9452 \\
        & Probabilistic & 0.9525 & 0.9526 \\
	\end{tabular}
 \captionsetup{width=13cm}
	\caption{Classification of the WDBC data set using CACTUS and SaNDA, showing consistency in performance. 
        In bold, the highest balanced accuracy is illustrated.}
	\label{tab:wdbc-classification}
\end{table}

\subsection{Thyroid data set}

For the Thyroid data set, both multiple and binary classification tasks have been applied.
For the binary version, we collapsed both hyper- and hypo-thyroidism in the class 1 by using 0 for the binarisation.
In this way, we considered healthy and thyroidal patients.
Since this data set has not been previously analysed, we applied several classic machine learning classifiers to obtain a comparison, as shown in Table~\ref{tab:thyroid-ml-classification}.

\begin{figure}[p]
	\centering
	\includegraphics[width=\textwidth]{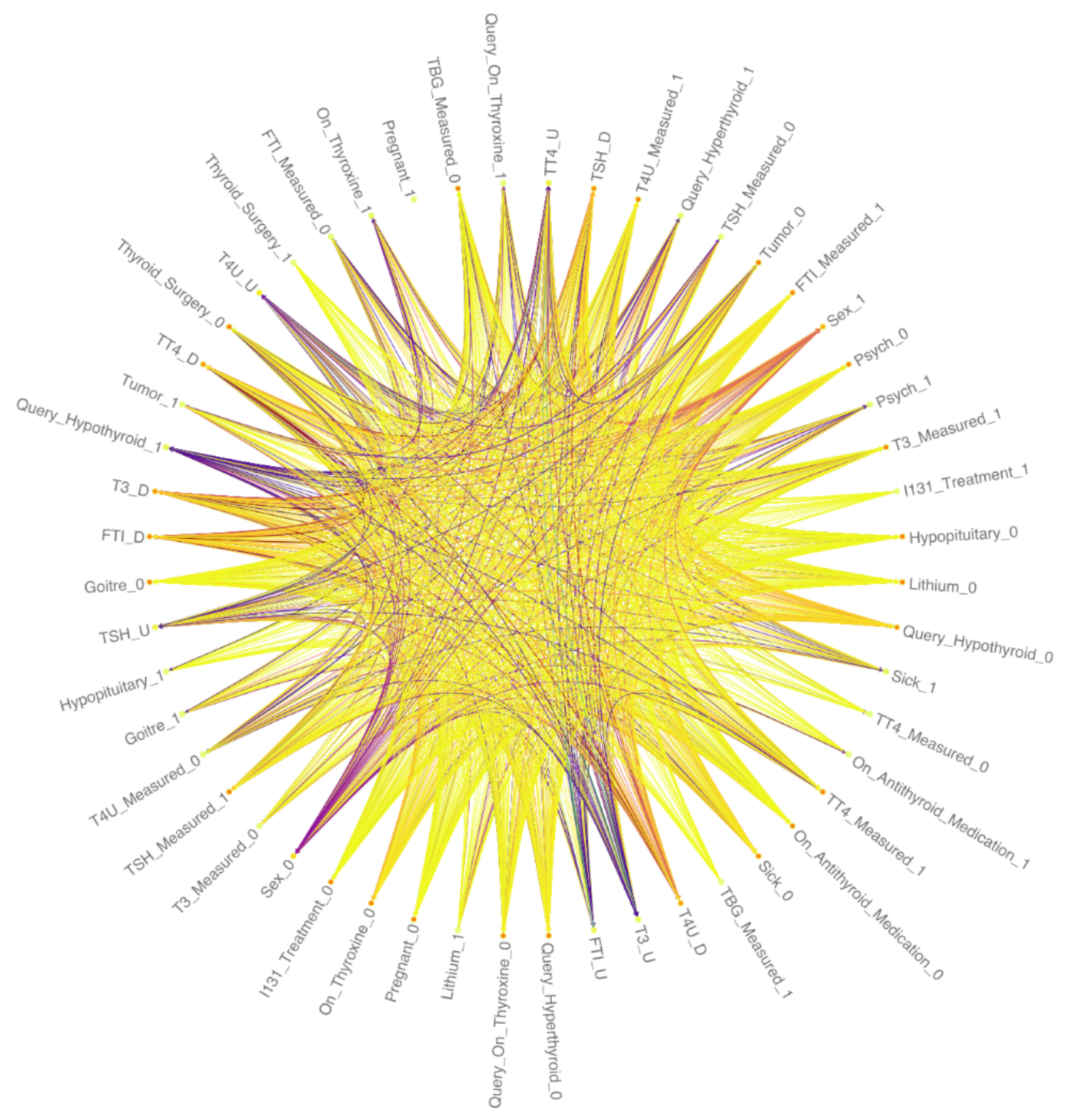}
	\includegraphics[width=.7\textwidth]{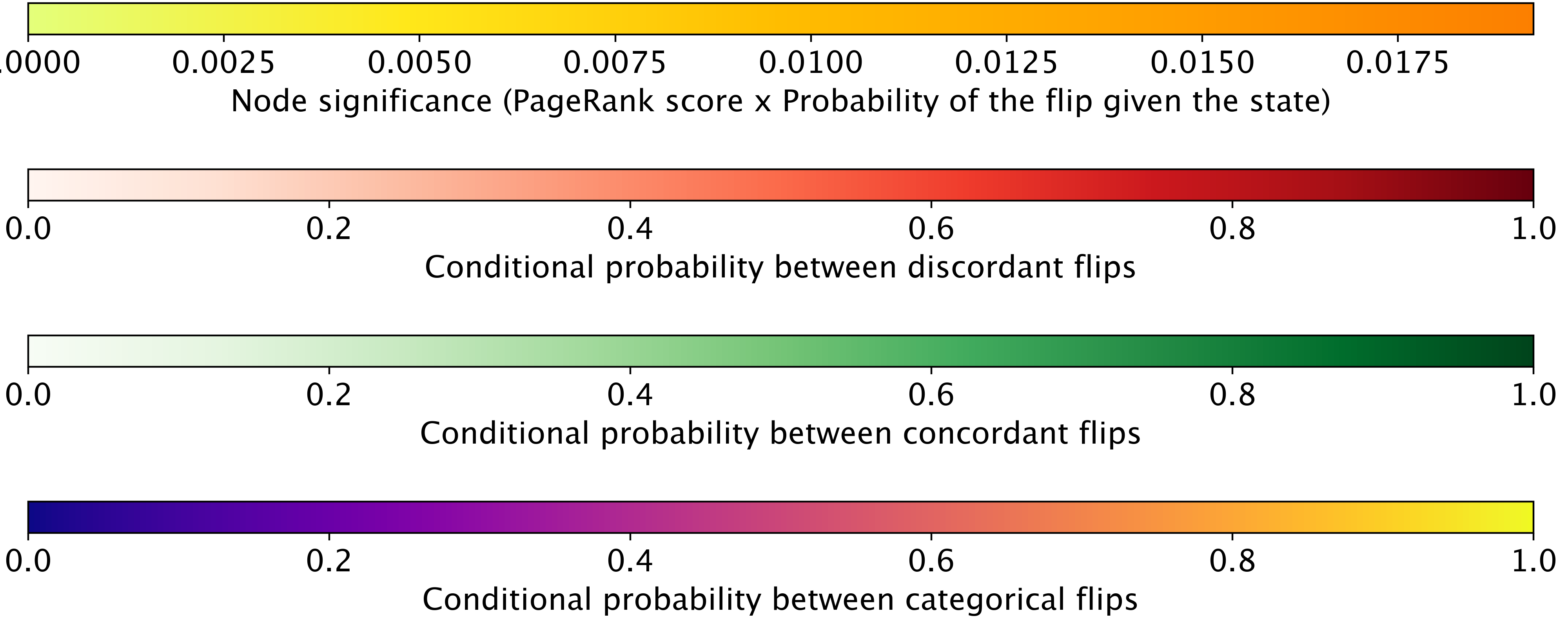}
	\caption{Knowledge graph representing the Healthy (0) class in the Thyroid data set.}
	\label{fig:thyroid-graph_0}
\end{figure}

\begin{figure}[p]
	\centering
	\includegraphics[width=\textwidth]{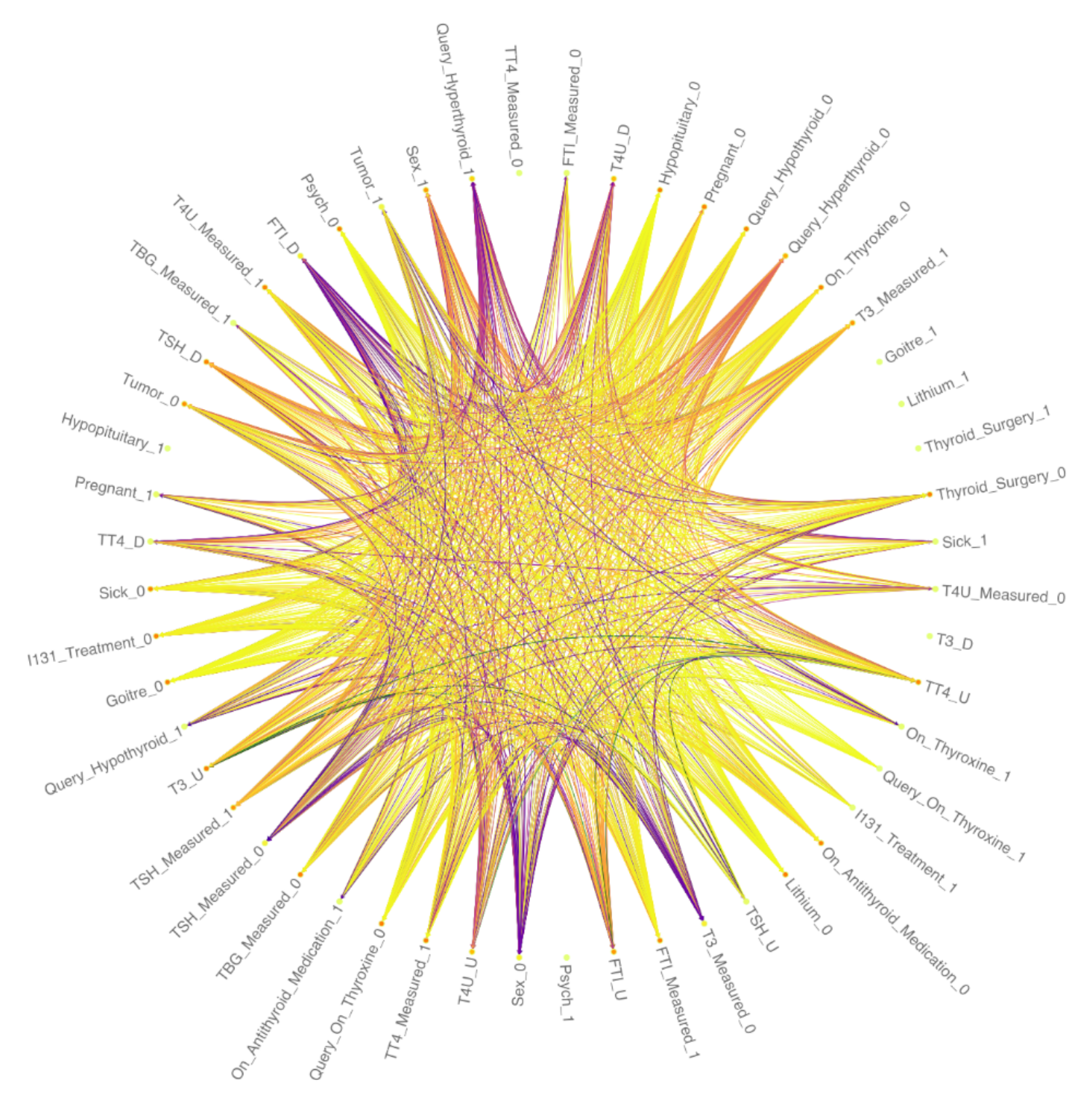}
	\includegraphics[width=.7\textwidth]{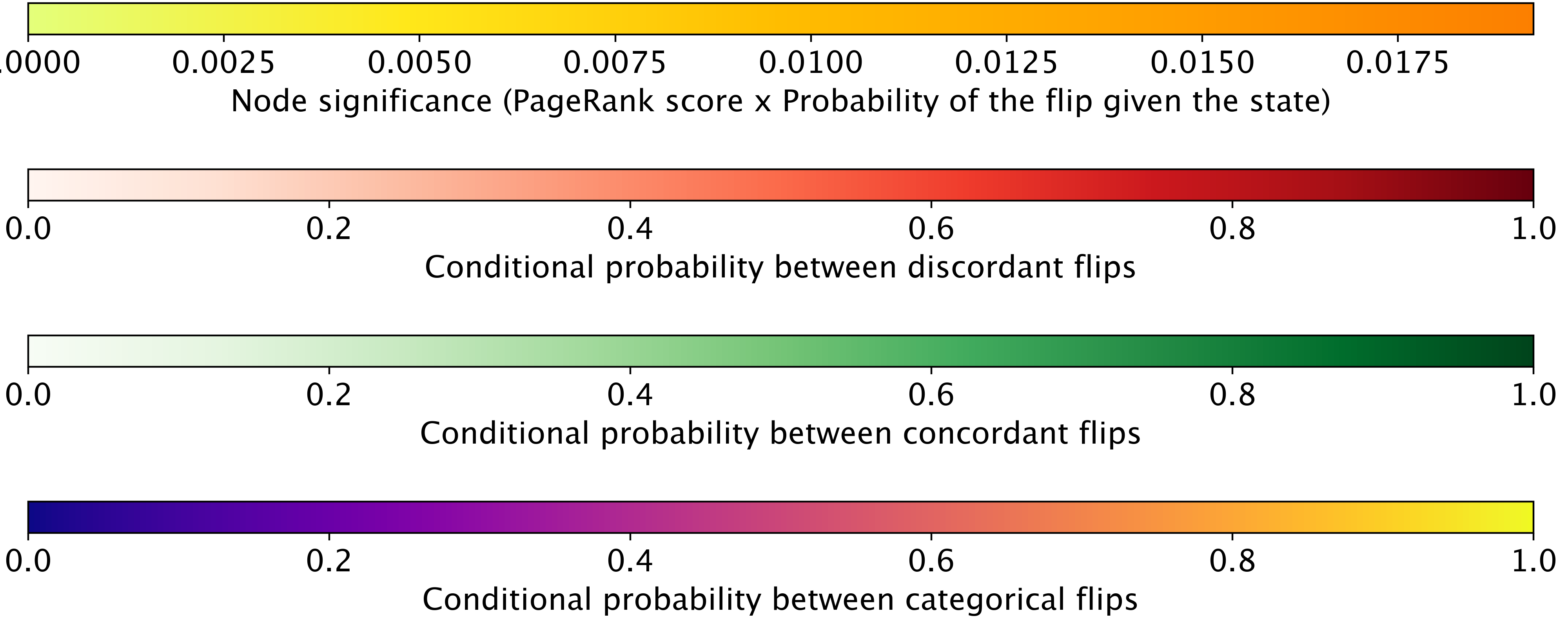}
	\caption{Knowledge graph representing the Hyperthyroidal (1) class in the Thyroid data set.}
	\label{fig:thyroid-graph_1}
\end{figure}

\begin{figure}[p]
	\centering
	\includegraphics[width=\textwidth]{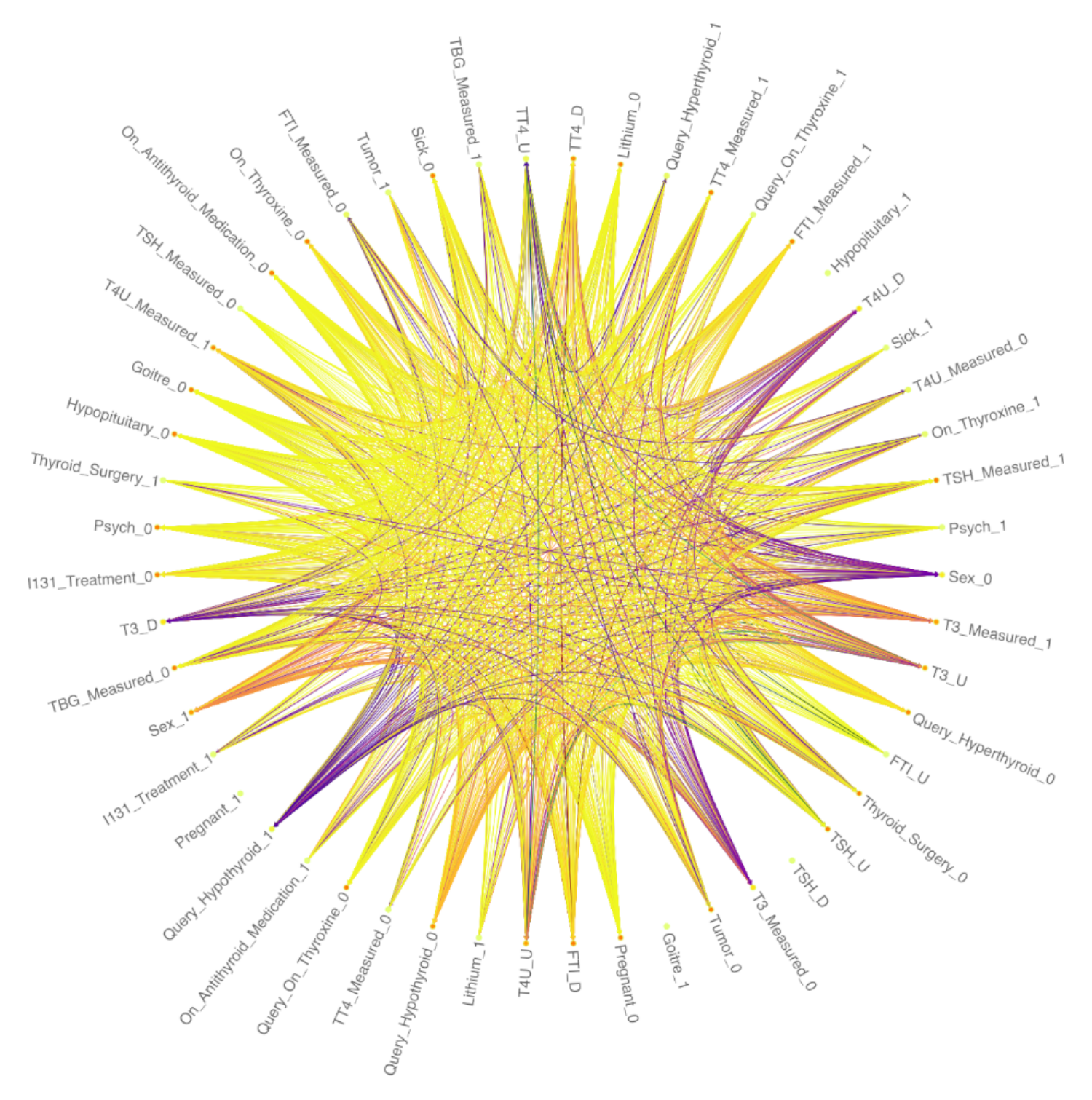}
	\includegraphics[width=.7\textwidth]{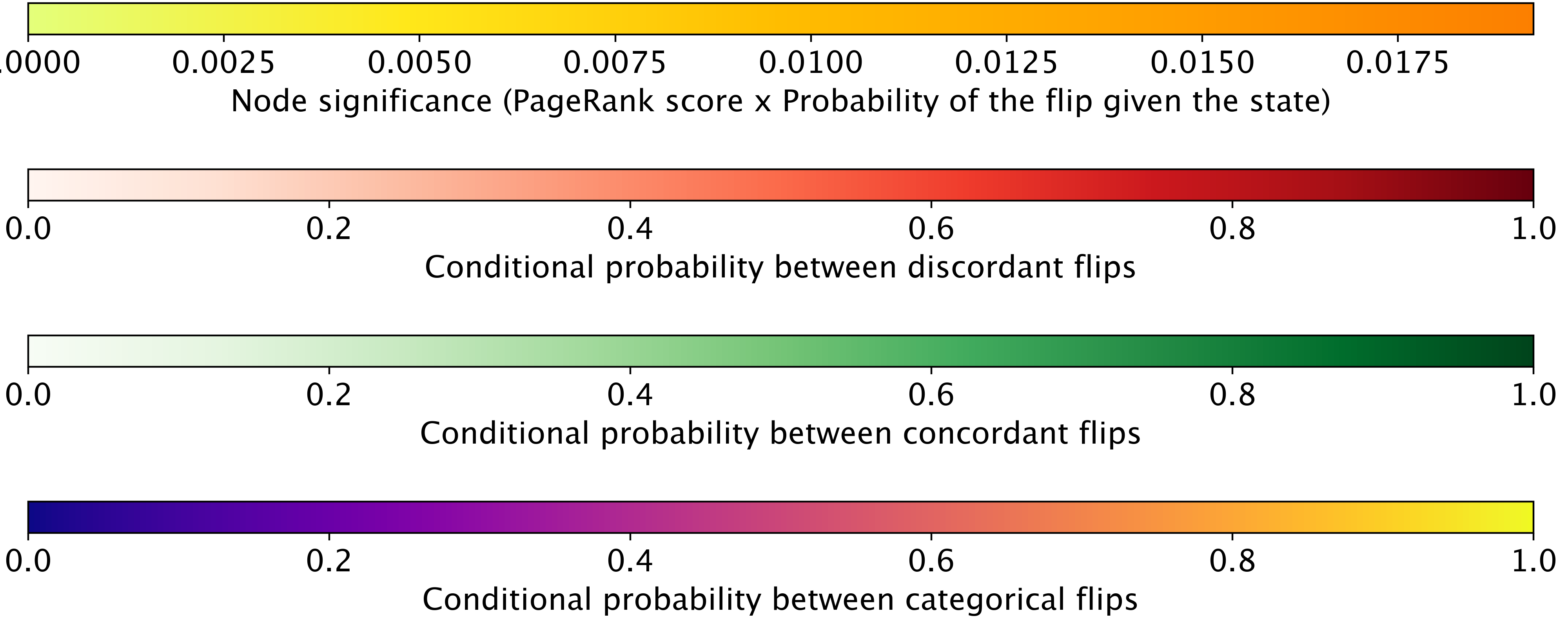}
	\caption{Knowledge graph representing the Hypothyroidal (2) class in the Thyroid data set.}
	\label{fig:thyroid-graph_2}
\end{figure}

\begin{figure}[p]
    \begin{minipage}{\linewidth}
    \centering
    \includegraphics[width=.5\textwidth]{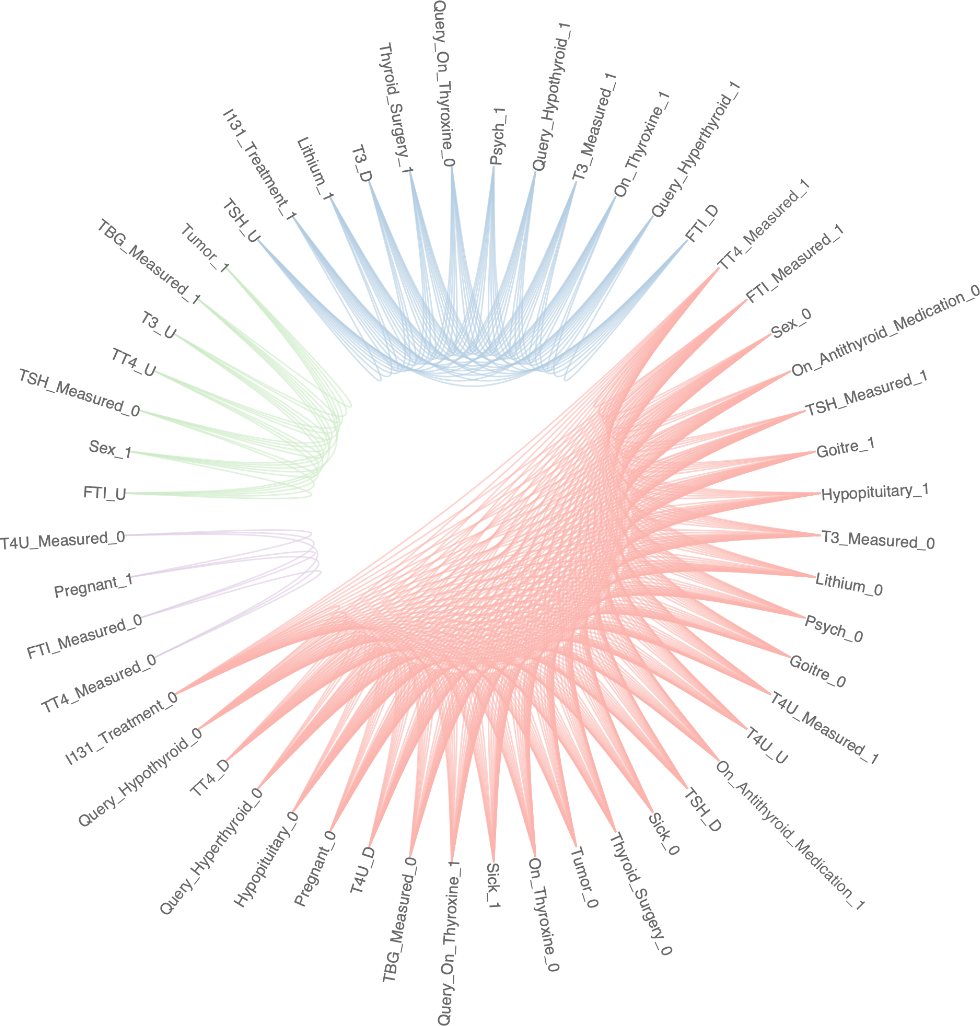}
    \end{minipage}
    \includegraphics[width=.5\textwidth]{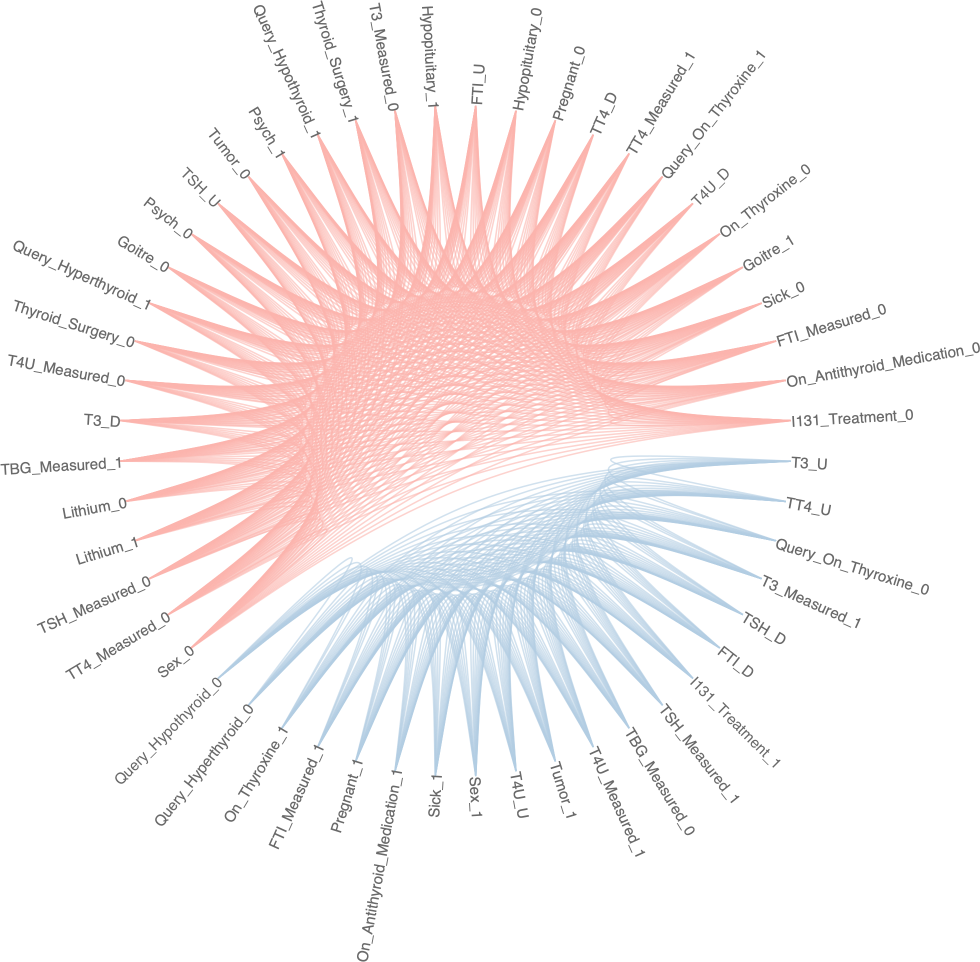}
    \includegraphics[width=.5\textwidth]{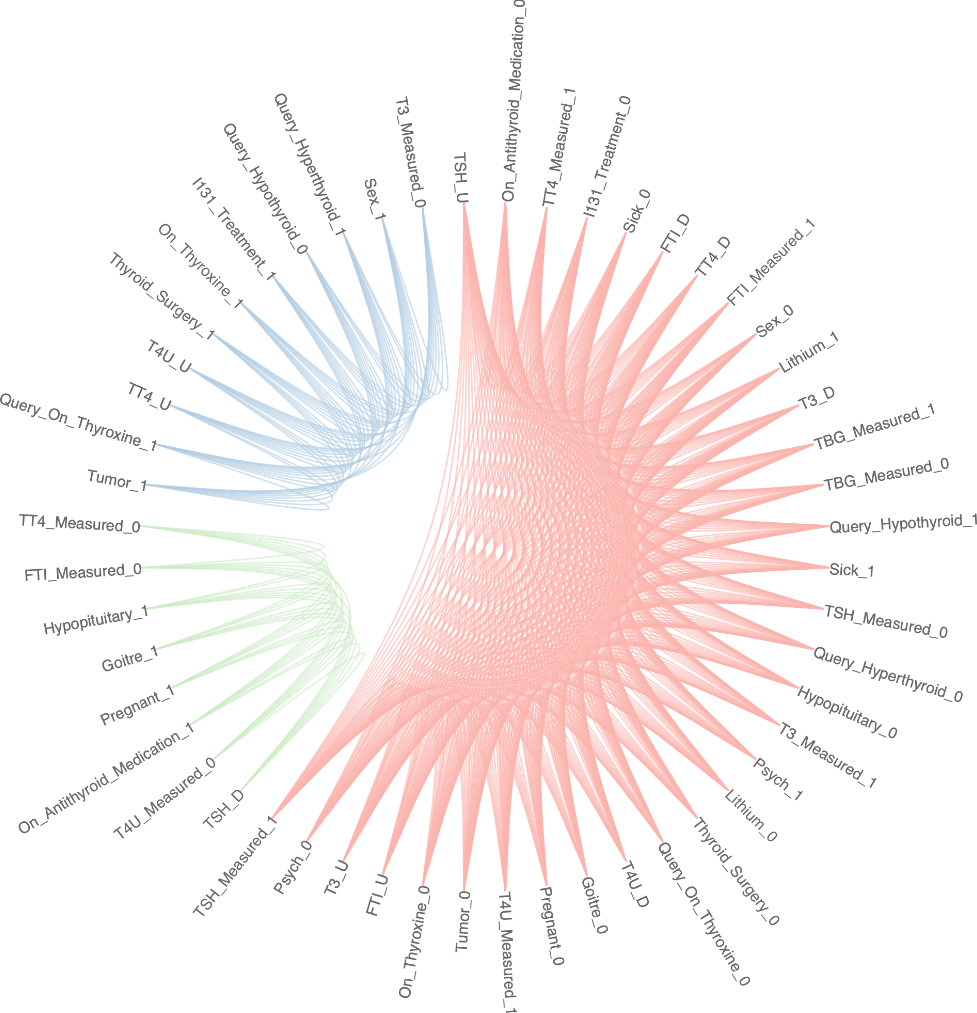}
	\caption{Communities created by the Greedy algorithm on the original thyroid data set on the healthy (top), hyperthyroidal (bottom left), and hypothyroidal (bottom right) classes. The red cluster in all three graphs has a subset of stable members that change ``polarity'' across classes, such as \emph{T4U}.}
	\label{fig:thyroid-greedy-communities}
\end{figure}

\begin{figure}[p]
    \begin{minipage}{\linewidth}   
    \centering
    \includegraphics[width=.5\textwidth]{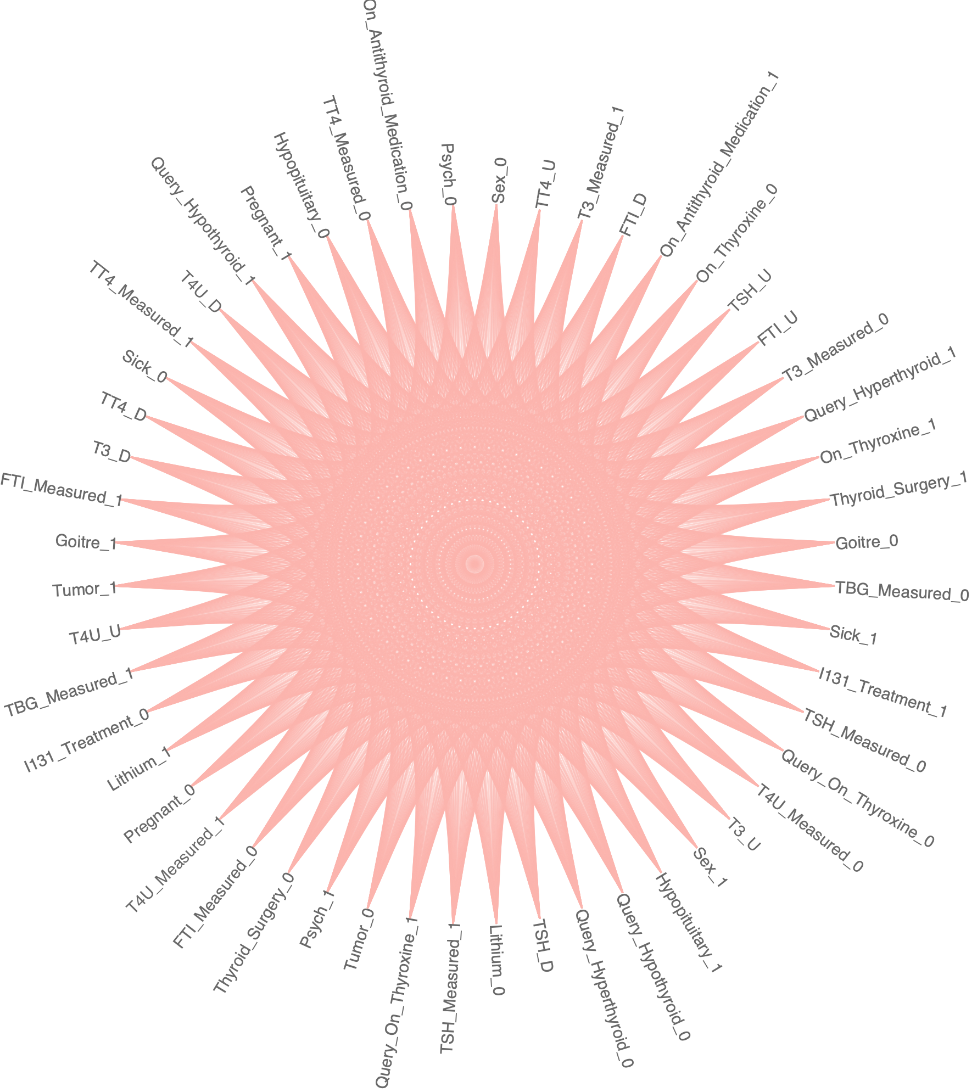}
    \end{minipage}
	\includegraphics[width=.5\textwidth]{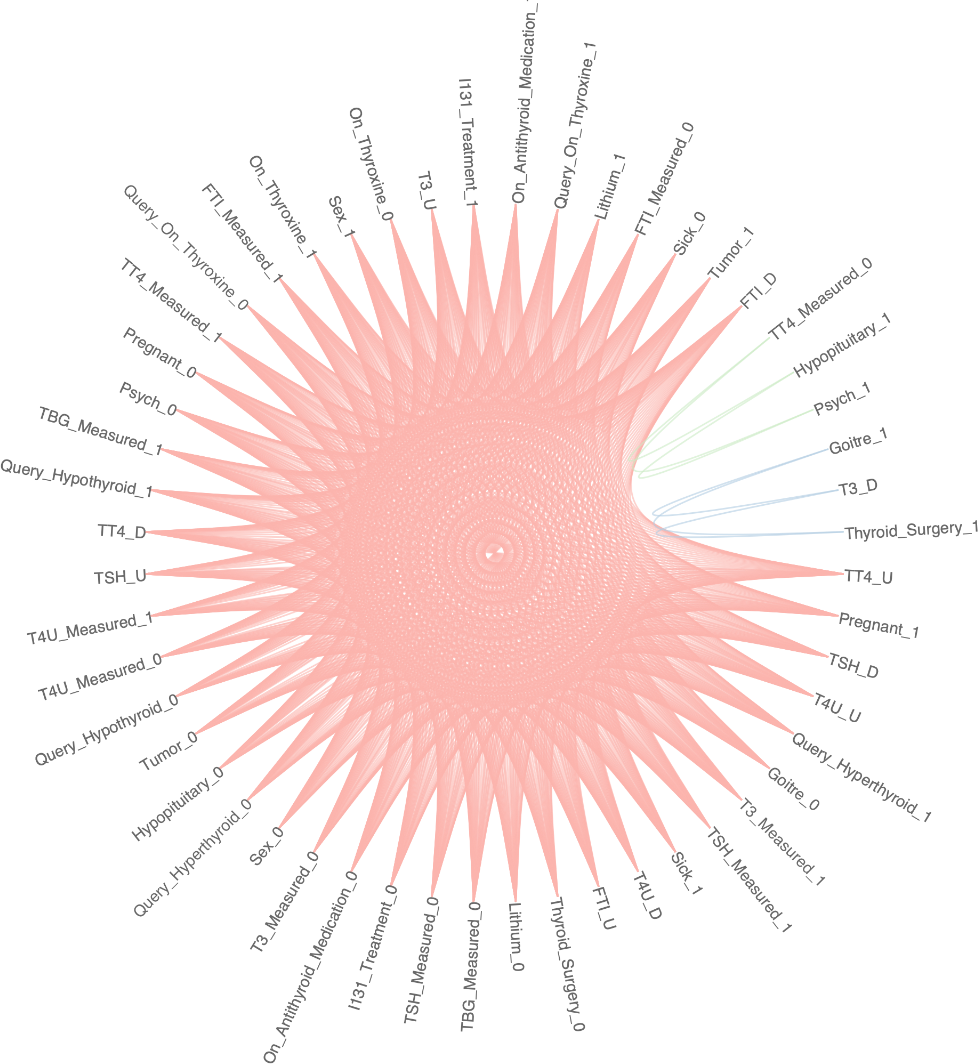}
	\includegraphics[width=.5\textwidth]{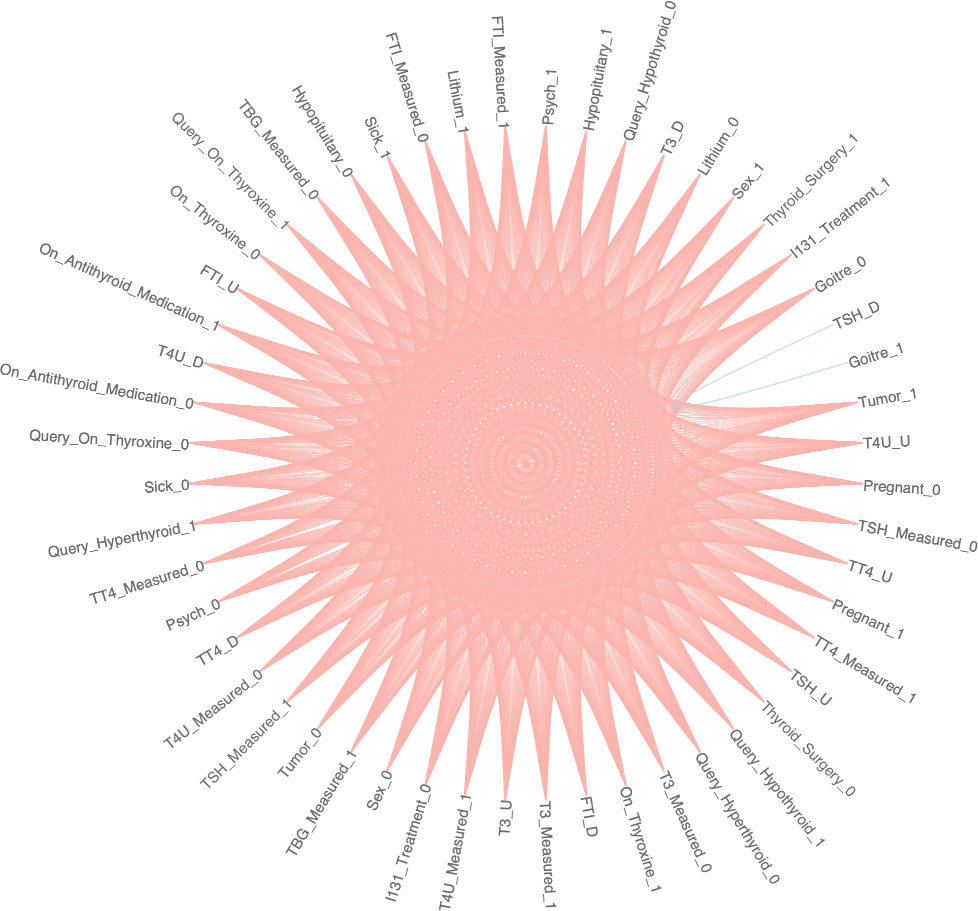}
	\caption{Communities created by the Label propagation algorithm on the original thyroid data set on the healthy (top), hyperthyroidal (bottom left), and hypothyroidal (bottom right) classes. The only stable difference shown between communities of healthy and thyroidal graphs is the \emph{Goitre\_1} flip, which relates with different flips in blue cluster of the hyper and hypo-thyroidal graphs.}
	\label{fig:thyroid-label-communities}
\end{figure}

\begin{figure}[p]
    \begin{minipage}{\linewidth}
    \centering
    \includegraphics[width=.5\textwidth]{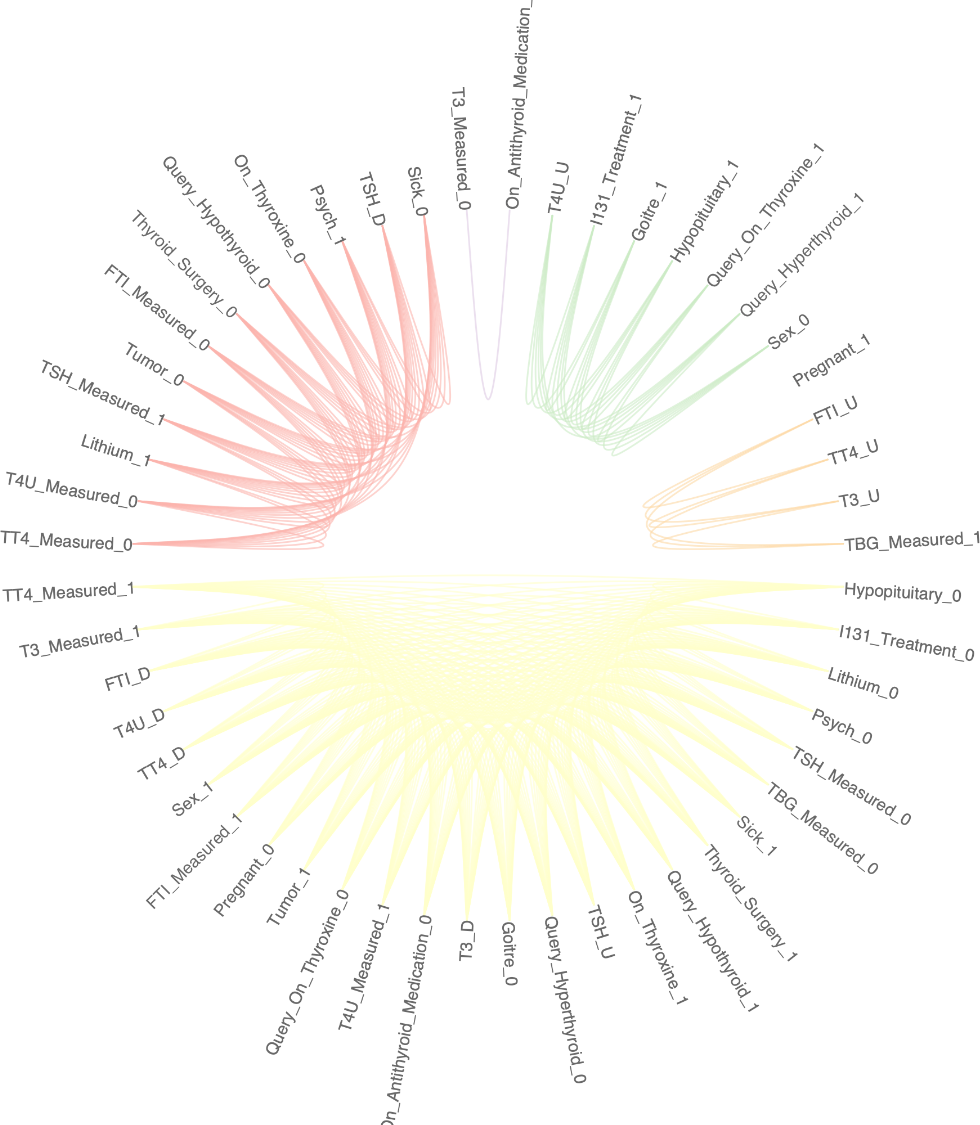}
    \end{minipage}
    \includegraphics[width=.5\textwidth]{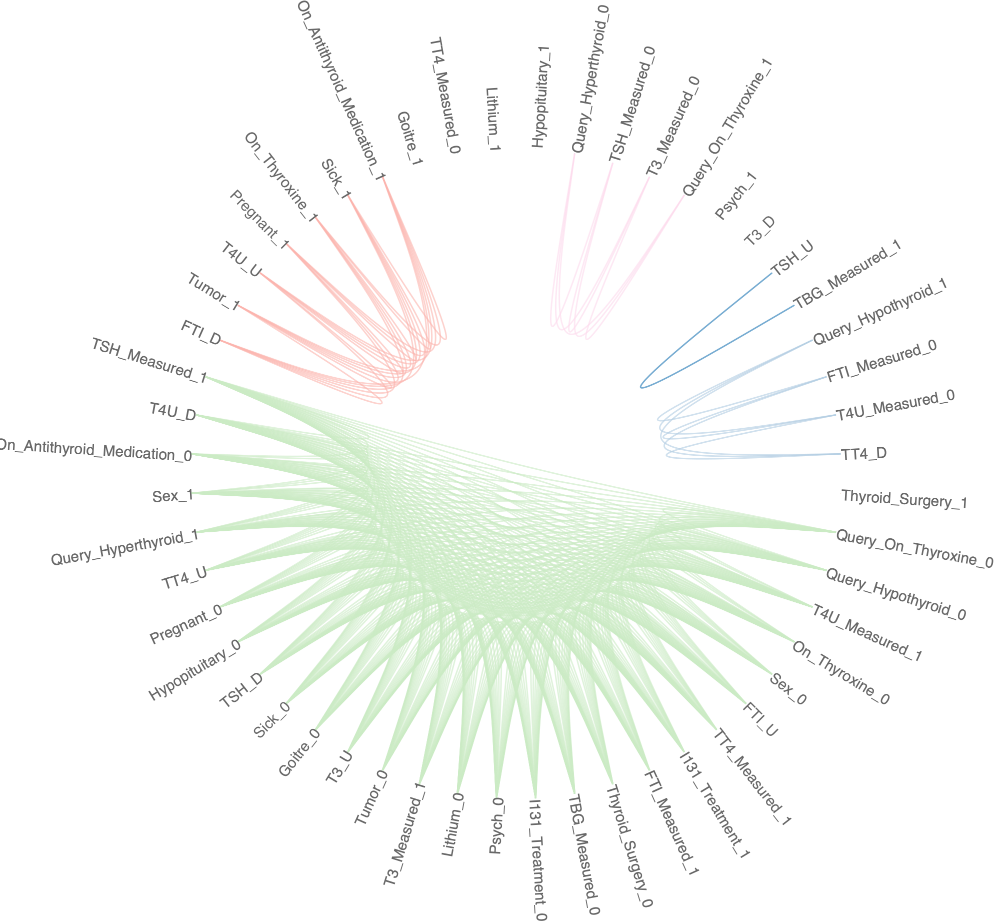}
    \includegraphics[width=.5\textwidth]
    {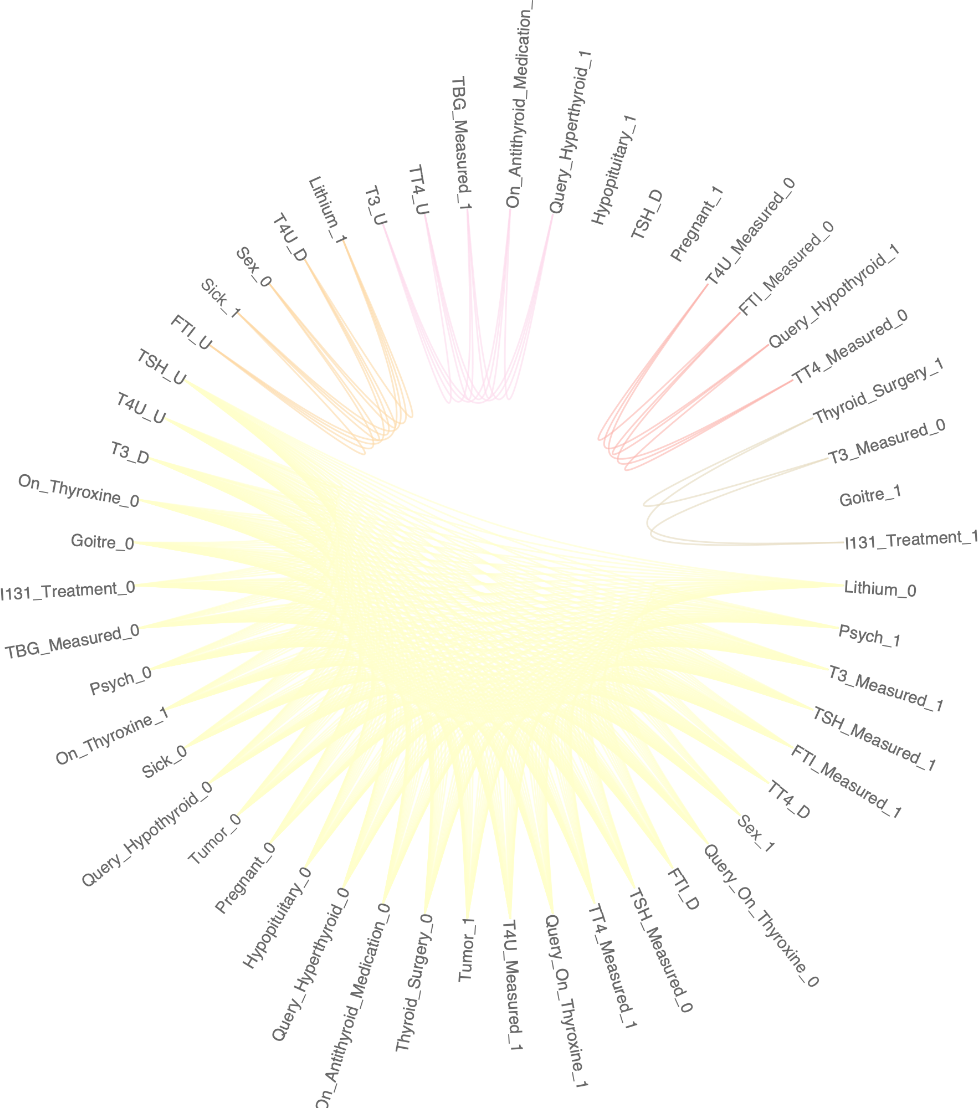}
    \caption{Communities created by the Louvain algorithm on the original Thyroid data set on the healthy (top), hyperthyroidal (bottom left), and hypothyroidal (bottom right) classes. Nodes without links have not been included in any community. The yellow communities in the upper and right graphs overlaps with the green community in the left.}
	\label{fig:thyroid-louvain-communities}
\end{figure}

\begin{figure}[p]
	\centering
	\includegraphics[width=\textwidth]{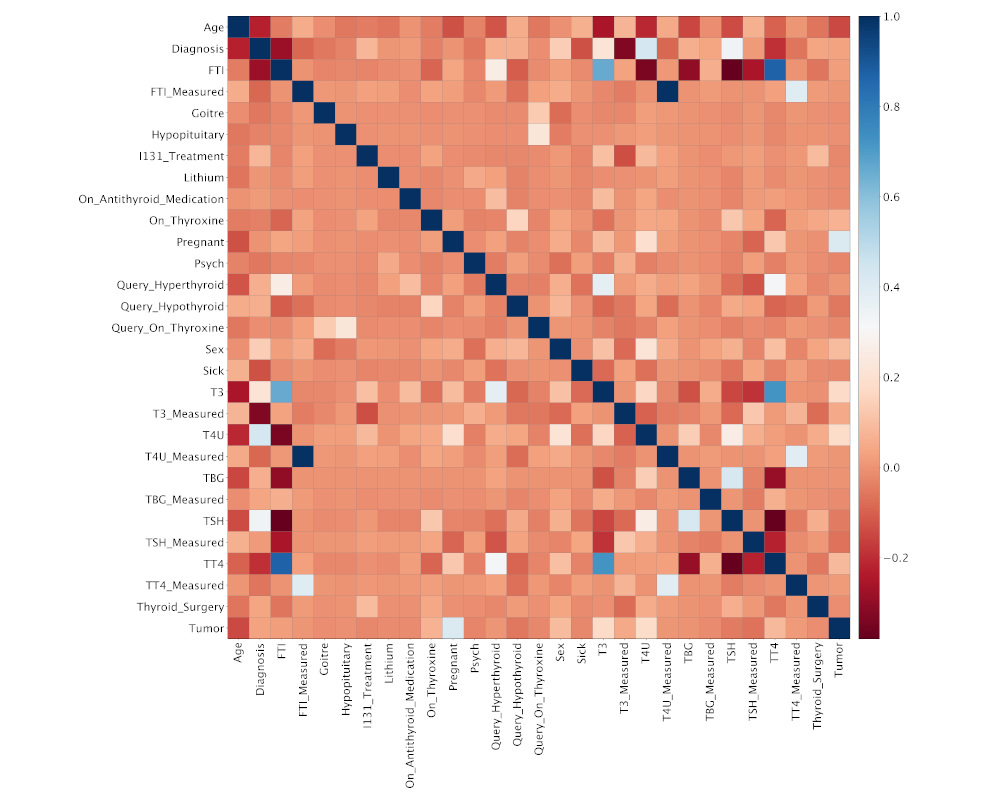}
	\caption{Correlation matrix on the Thyroid data set.}
	\label{fig:thyroid-correlation}
\end{figure}

Figures~\ref{fig:thyroid-graph_0},~\ref{fig:thyroid-graph_1}, and~\ref{fig:thyroid-graph_2} show how the connections change across healthy, hypothyroidism and hyperthyroidism, respectively, by seeing the difference in nodes with the highest significance (i.e., \emph{TT4\_U}).
This is also visible in the communities created by the Greedy, Label propagation, and Louvain algorithms, showed respectively in Figures~\ref{fig:thyroid-greedy-communities},~\ref{fig:thyroid-label-communities}, and~\ref{fig:thyroid-louvain-communities}. 
In Figure~\ref{fig:thyroid-correlation}, the correlation patterns between the attributes are shown, in particular between \emph{Diagnosis} and \emph{T3\_Measured}, \emph{FTI}, and \emph{TT4} (anti-correlations), as discussed in the Discussion Section.

\begin{table}[htb]
	\begin{tabular}{c c | c c}
		Binarisation & Target Divider & Model         & Balanced Accuracy   \\ \hline
		Original     & $0$              & PageRank    & \textbf{0.9323}            \\
		&                & Probabilistic & \textbf{0.9323}   \\ \hline
		& $1$              & PageRank    & $0.9144$            \\
		&                & Probabilistic & $0.9144$            \\ \hline \hline
		$0$          & $0$              & PageRank    & $0.9108$            \\
		&                & Probabilistic & $0.9108$            \\
	\end{tabular}
  \captionsetup{width=13cm}
	\caption{Classification of the Thyroid data set through CACTUS. In bold, the highest balanced accuracy achieved is illustrated. In this scenario, the \emph{Original} binarisation preserved the three considered classes, while $0$ in binarisation partitioned the data set in healthy and thyroidal subjects.}
	\label{tab:thyroid-classification}
\end{table}

\begin{table}
	\begin{tabular}{c | c}
		Model & Balanced Accuracy \\ \hline
		CACTUS (Probabilistic) & $0.954$ \\
		Ridge classifier & $0.85$ \\
		Logistic regression classifier & $0.92$ \\
		Support Vector Machine & $0.8988$ \\
		Stochastic Gradient Descent Classifier & $0.9228$ \\
		Random Forest Classifier & \textbf{$0.9949$}
	\end{tabular}
  \captionsetup{width=13cm}
	\caption{Comparison between Probabilistic (divider 0) and standard machine learning classifiers on the three-classes Thyroid data set. CACTUS proves better performance than many of them except for the Random Forest. Notably, Random Forest is the less interpretable in the listed models, and CACTUS is the best alternative for having both classification performance and transparency.}
	\label{tab:thyroid-ml-classification}
\end{table}

\section{Discussion}

\subsection{WDBC data set}
\begin{sidewaysfigure}[p]
	\centering
	\includegraphics[width=\textwidth]{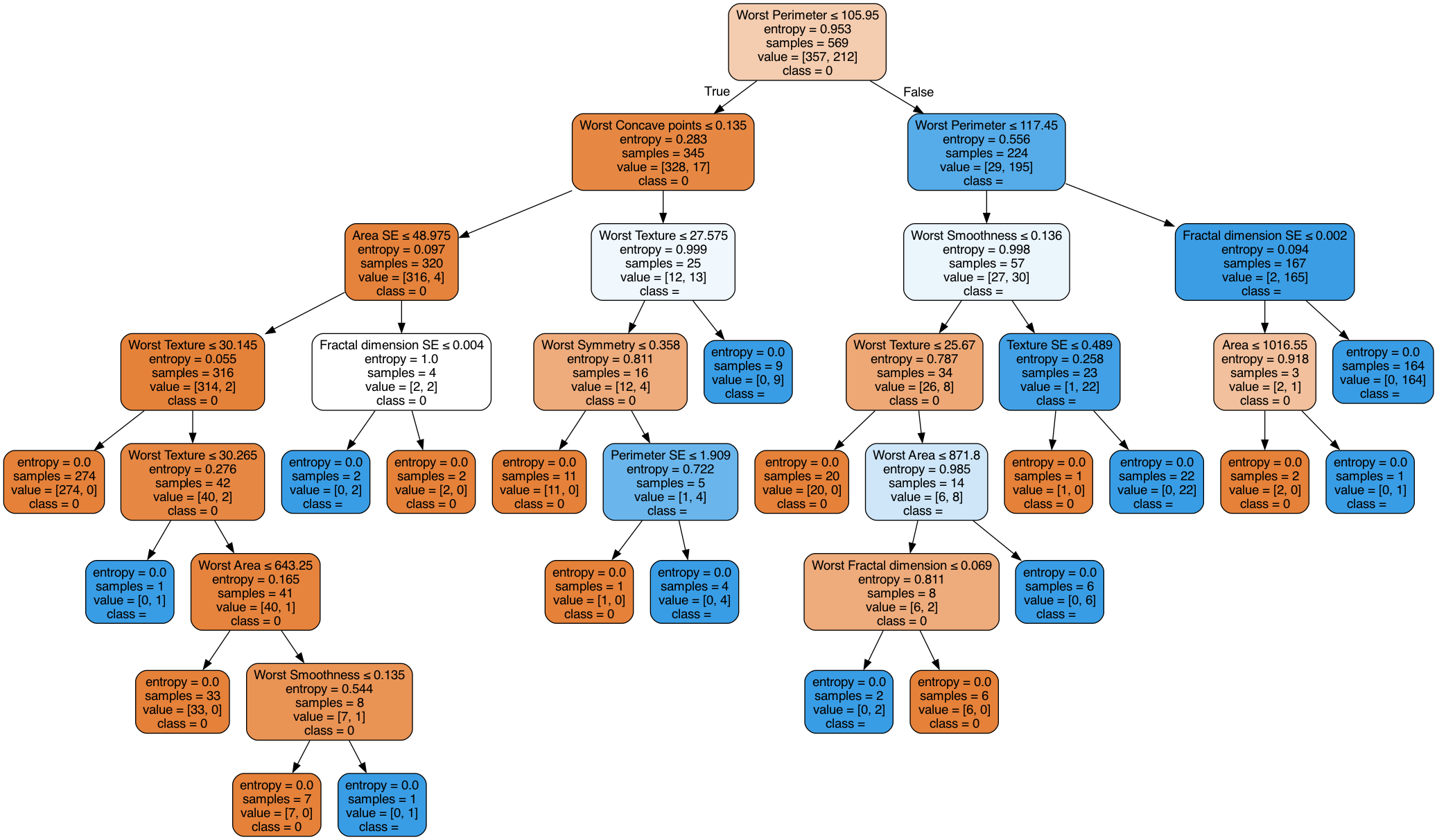}
	\caption{Binary decision tree on the WDBC data set. Each bifurcation represents a linear separation operated on one of the attributes of the data set (i.e. Worst Perimeter $\leq 117.45$). Orange leaves represent pure partitions of benign breast cancers and blue malignant cancers.}
	\label{fig:wdbc-decision-tree}
\end{sidewaysfigure}

Classifying the kind of breast cancer in the WDBC data set using CACTUS allows the user to visualise the most influential elements in the decision process.
In Figure~\ref{fig:wdbc-decision-tree}, the auxiliary decision tree shows the number of linear separations, chosen using the entropy criterion, to obtain pure classes (coloured leaves), where the first splits have been on \emph{Worst perimeter}, \emph{Worst concave points}, \emph{Worst texture}, \emph{Worst smoothness}, \emph{Area SE}, and \emph{Fractal dimension SE}.
The number of linear separations shows the complexity required to obtain pure partitions, despite having only three classes.

\begin{figure}[p]
	\centering
	\includegraphics[width=1.1\textwidth]{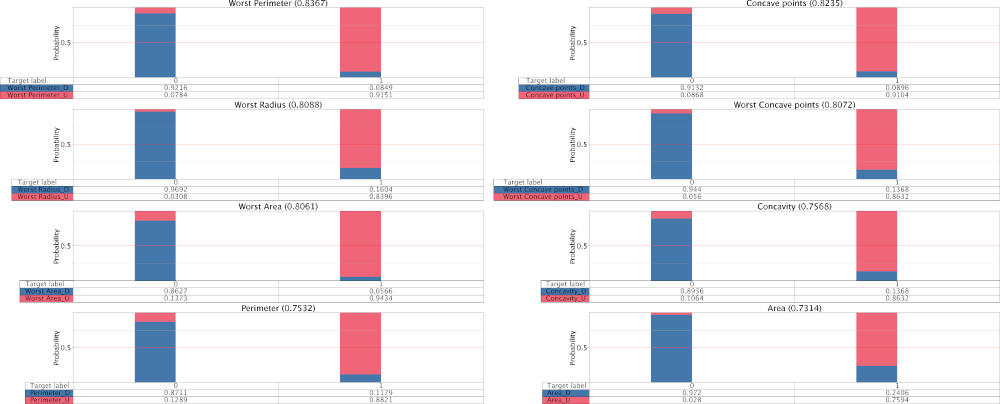}
	\caption{Transitions in the probability of each marker in the WDBC across benign (0) and malignant (1) cancer. The title of each plot contains the average rank $\Bar{R_x}$ computed through Equation \eqref{eq-avg-rank}.}
	\label{fig:wdbc-all-node-significance}
\end{figure}

CACTUS shares with the binary decision tree the attributes \emph{Worst perimeter} and \emph{Worst concave points} as the most meaningful for splitting the data set between the two classes, as shown in Figure~\ref{fig:wdbc-all-node-significance}.
Intuitively, the more that the probability distribution of the flips of a marker changes, the more useful that the marker is for assigning a patient to one of the two classes, as represented by the average rank $\Bar{R_x}$ of a marker $x$ computed in Equation \eqref{eq-avg-rank}.
This is reflected in the knowledge graphs, were a flip might be much more influential than its twin in a graph and surrender to it in the next one, causing very different scores when the classification function is applied on the knowledge graphs.
For example, \emph{Smoothness} (not shown for space constraints) has a very balanced distribution for the benign cancer ($0.52$ Up and $0.48$ Down) and a more skewed distribution ($0.86$ Up and $0.14$ Down) for the malignant breast cancer (where malignant cancers are generally not very smooth), and therefore its average rank $\Bar{R_x}$ is not outstanding ($0.3394$).
Conversely, \emph{Worst perimeter} is much more informative due to the inverted distributions in the benign and the malignant types, and therefore gets a very high $\Bar{R_x}$ ($0.8367$).

The differences in the connections and the communities across the knowledge graphs is based and justified by Figure~\ref{fig:wdbc-all-node-significance}, where the differences in the distributions drive such changes to appear by changing the centrality of each flip.

Comparing the strongest correlations of the Diagnosis column with the distributions in Figure~\ref{fig:wdbc-all-node-significance}, we can observe that \emph{Area}, \emph{Concave points}, \emph{Concavity}, and \emph{Perimeter} have the most noticeable changes in their flips distributions, while other markers shown, such as \emph{Smoothness}, have both a less impressive change in the distributions and no significant correlation with the \emph{Diagnosis column}.
This is useful to confirm that the markers used by CACTUS to operate the classification task are actually bound to the label to predict, and therefore the building of the knowledge graph can be validated and supported.
Other well correlated variables, namely \emph{Worst concave points}, \emph{Worst area}, \emph{Worst perimeter}, \emph{Worst radius}, and \emph{Radius}, show important differences in their distributions as well, but have been omitted in Figure~\ref{fig:wdbc-all-node-significance} for space constraints.

\subsection{Thyroid data set}

\begin{figure}[p]
\centering
	\includegraphics[width=\textwidth]{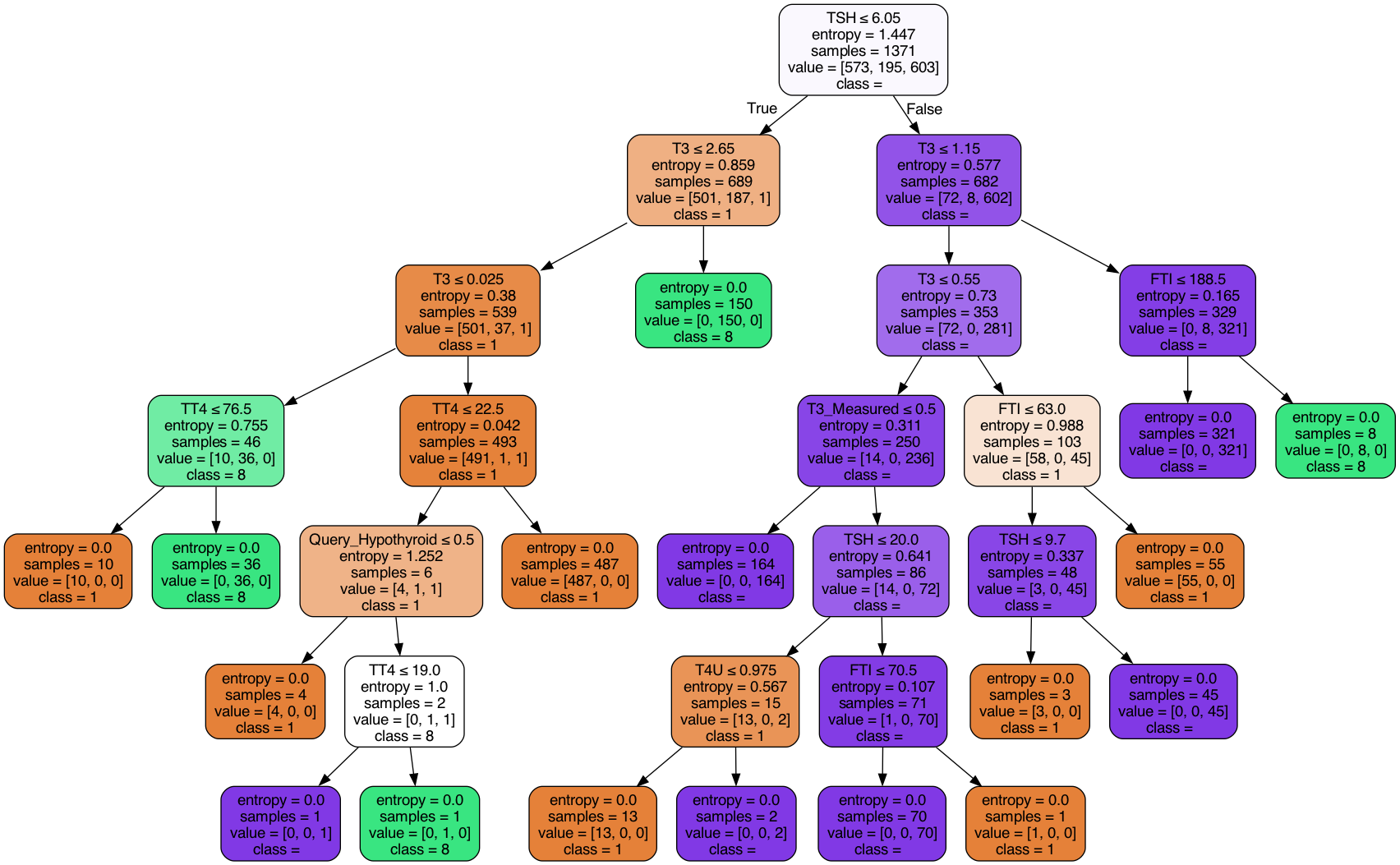}
	\includegraphics[width=\textwidth]{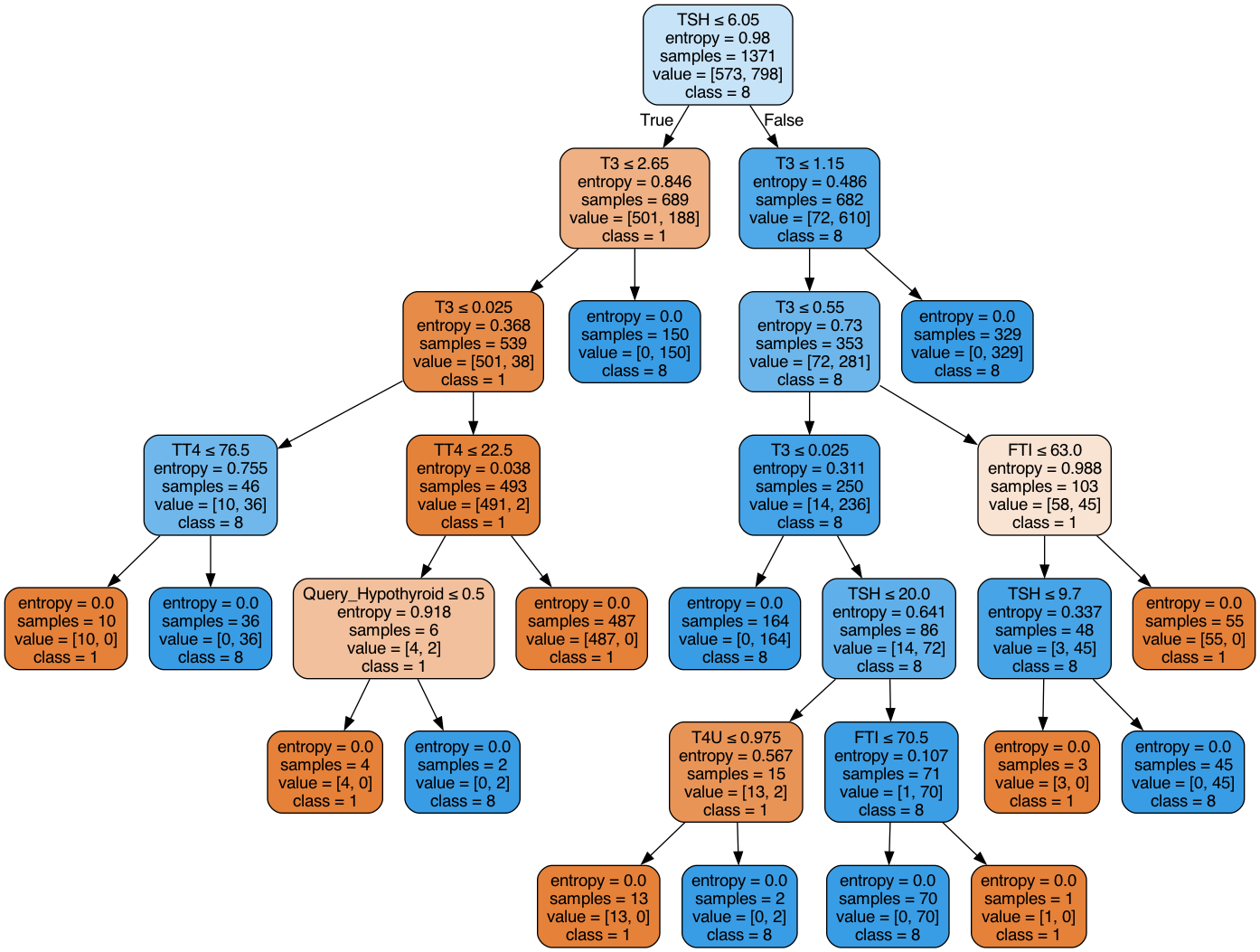}
	\caption{Binary decision tree computed on the Thyroid data sets for the original three classes (left) and binarisation of them (right).}
	\label{fig:thyroid-decision-tree}
\end{figure}

\begin{figure}[p]
	\centering
	\includegraphics[width=1.1\textwidth]{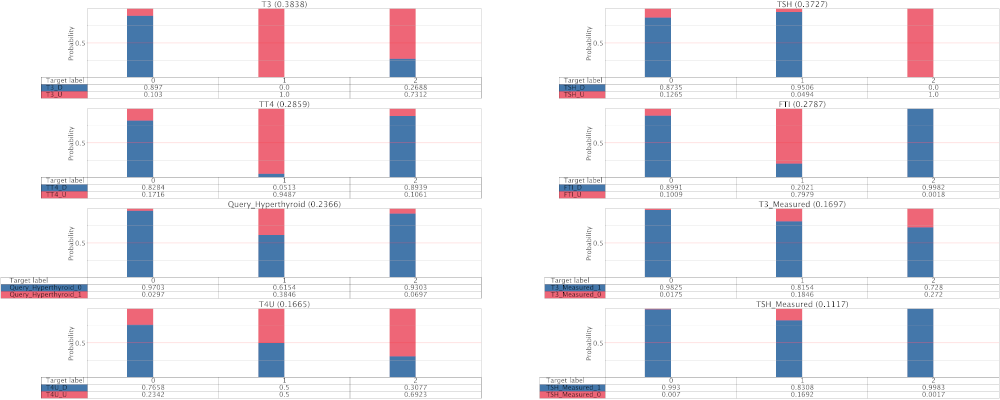}
	\caption{Changes in the probability distributions of the most influential flips of the Thyroid data set across the Healthy (0), Hyperthyroidal (1), and Hypothyroidal (2) classes. The title of each plot contains the average rank $\Bar{R_x}$ computed through Equation \eqref{eq-avg-rank}.}
	\label{fig:thyroid-all-node-significance}
\end{figure}

The classification of the Thyroid data set tested the capacity of CACTUS of dealing with multiple classification tasks.
Table~\ref{tab:thyroid-classification} shows the unedited performance achieved by CACTUS on both original and binarised data set, as SaNDA does not support categorical variables.
Table~\ref{tab:thyroid-ml-classification} shows that CACTUS has very good performance compared to standard machine learning models, second only to the Random Forest classifier.
The training/test splits for these traditional classifiers contained 90$\%$ and 10$\%$ of the data, respectively.
Notably, Random Forest is the less interpretable of the chosen models, and the trade-off between interpretability and performance is playing a central role in driving the transition towards more transparent models for critical sectors.
Figure~\ref{fig:thyroid-decision-tree} shows the linear separations to obtain pure partitions of both original and binarised data set.
Interestingly, they are not very different visually, showing that the re-partitioning the data set is preserving the better part of the difference between healthy and thyroidal partitions.
Both trees perform the first splits on \emph{TSH}, \emph{T3}, \emph{FTI}, and \emph{TT4}.
Considering Figure~\ref{fig:thyroid-all-node-significance}, the same markers show significant changes in their distributions across the three original classes, and \emph{Sex\_1} (Female sex) is statistically more likely to have thyroid-related pathologies (data not shown due to space constraints).
Most of the influential markers are specific to a particular thyroidal condition, such as \emph{TT4} and \emph{FTI}, but others seem to be more common between hyper- and hypo-thyroidism, such as \emph{T4U} and \emph{T3}, despite they still show a preference for hypothyroidism and hyperthyroidism, respectively.
Interestingly, the Greedy algorithm shows more communities in the healthy class than in the others, but Label propagation and Louvain do the opposite, creating more communities in the thyroidal classes than in the healthy one.

In Figure~\ref{fig:thyroid-correlation}, we can describe how \emph{Diagnosis} correlates with the other attributes, and it is interesting to observe how the preferred attributes for both CACTUS and the binary decision tree, namely \emph{TSH}, \emph{T3}, \emph{FTI}, and \emph{TT4}, have no or even negative correlation.

\section{Conclusions}
The new architecture presented in this work has been shown it can perform different kind of classification tasks with a high degree of flexibility and transparency.
It automatically provides several internal representations to the user to describe its functioning and what elements are considered fundamental to assign a specific class rather than another.
The inclusion of categorical features is an important addition to allow the full use and interpretation of these variables, especially for medical data where medications, genetic, or habitual variables are included.
The integration of external methods, namely binary decision trees and correlation matrices, give a reference against which comparing the knowledge graph compositions.
Using abstractions for representing information is a novel avenue yet to be explored, and future directions regard exploiting it in different domains, as a fundamental block for models different from the current panorama in AI.

\bibliography{bibliography}
\newpage

\end{document}